\documentclass{article}

\usepackage{microtype}
\usepackage{graphicx}
\usepackage{subfigure}
\usepackage{multirow}
\usepackage{threeparttable}
\usepackage{makecell}

\usepackage{booktabs} 

\usepackage{hyperref}

\usepackage{amsfonts,amsmath,amssymb,amsthm,mathtools} 
\usepackage{wrapfig}  
\usepackage{mdwlist}  
\usepackage{sidecap}  
\usepackage{bbm}
\usepackage{bm,amsbsy} 
\newtheorem{theorem}{Theorem}

\usepackage[accepted]{icml2019}

\icmltitlerunning{Radial and Directional Posteriors for Bayesian Neural Networks}

\newcommand{\punt}[1]{}

\newtheorem{prop}{Proposition}

\newcommand{\reals}{\mathbb{R}}

\newcommand{\vw}{\mathbf{w}}
\newcommand{\vx}{\mathbf{x}}

\newcommand{\bq}{\begin{equation}}
\newcommand{\eq}{\end{equation}}
\newcommand{\ba}{\begin{eqnarray}}
\newcommand{\ea}{\end{eqnarray}}

\def\R{{\reals}}

\newcommand{\remove}[1]{}

\newcommand{\Nrm}{\mathcal{N}}
\newcommand{\Em}{\mathbb{E}}

\renewcommand{\eqref}[1]{eq.~\ref{eq:#1}}

\newcommand{\figref}[1]{Fig.~\ref{fig:#1}}  
\newcommand{\secref}[1]{Sec.~\ref{sec:#1}}  
\newcommand{\subsecref}[1]{Sec.~\ref{subsec:#1}}  
\newcommand{\suppsecref}[1]{Appendix Sec.~\ref{supp:#1}}  

\newcommand{\Dat}{\mathcal{D}}

\newcommand{\vphi}{\mathbf{\ensuremath{\bm{\phi}}}}

\newcommand{\vmu}{\mathbf{\ensuremath{\bm{\mu}}}}
\newcommand{\vtheta}{\mathbf{\ensuremath{\bm{\theta}}}}

\newcommand{\mW}{\mathbf{W}}

\usepackage{color}  
\definecolor{gray}{rgb}{0.5, .5, .5}

\begin{document}

\twocolumn[
\icmltitle{Radial and Directional Posteriors for Bayesian Neural Networks}



\icmlsetsymbol{intern}{*}

\begin{icmlauthorlist}
\icmlauthor{Changyong Oh}{uva,intern}
\icmlauthor{Kamil Adamczewski}{mpi}
\icmlauthor{Mijung Park}{mpi}
\end{icmlauthorlist}

\icmlaffiliation{uva}{QUvA lab, Informatics Institute, University of Amsterdam, Amsterdam, Netherlands}
\icmlaffiliation{mpi}{Max Planck Institute for Intelligent Systems, T{\"u}bingen, Germany}

\icmlcorrespondingauthor{Changyong Oh}{c.oh@uva.nl}

\icmlkeywords{Variational Inference, Bayesian Neural Networks, High Dimension, von-Mises Fisher Distribution, Compression}

\vskip 0.3in
]



\printAffiliationsAndNotice{\icmlInternship}

\begin{abstract}
\vspace{1em}

We propose a new variational family for Bayesian neural networks. 
We decompose the variational posterior into two components, where
the {\it{radial}} component captures the strength of each neuron in terms of its magnitude; while the {\it{directional}} component captures the statistical dependencies among the weight parameters. %
The dependencies learned via the directional density provide better modeling performance compared to the widely-used Gaussian mean-field-type variational family.
In addition, the strength of input and output neurons learned via the radial density provides a structured way to compress neural networks. 
Indeed, experiments show that our variational family improves predictive performance and yields compressed networks simultaneously.


\end{abstract}

\vspace{1em}
\section{Introduction}\label{sec:intro}

Neural networks have recently become revolutionary tools to solve numerous statistical problems in science and industry.
However, the uncertainty of the network weight estimates is often neglected, although 
there is a growing necessity to model the uncertainty in many application domains such as medical decision-making and climate prediction~\cite{slingo2011uncertainty}.   

Reasoning about uncertainties in the neural network models was initiated by a few seminal papers \cite{neal2012bayesian, mackay1995probable, dayan1996varieties}.
These works aimed to develop Bayesian methods for neural network models, suggesting a new research direction, called {\it{Bayesian neural networks}} (BNNs).  
Recent attempts to develop such techniques include \cite{ blundell2015weight, hernandez2015probabilistic, gal2016dropout, louizos2016structured} among many.
%
These, however, rely on prior and posterior pairs that are often chosen for convenience in inference, namely, computational tractability. The so-called {\it{mean-field}} variational family in these works assumes the posterior distributions to be all factorizing, and hence neglects the possibility of modelling statistical dependencies  (i.e., correlations) among weight parameters \cite{graves2011practical, blundell2015weight, kingma2015variational, neklyudov2017structured, ullrich2017soft, molchanov2017variational}.   

Capturing dependencies between the weight parameters and their uncertainties is likely to yield better models in terms of predictability. For instance, \citet{louizos2016structured, pmlr-v54-sun17b} propose Gaussian posteriors with covariance matrices of certain structures such that each covariance can capture the dependencies among the input and output dimensions of each layer.
Rather than proposing a new variational family, \citet{NIPS2017_7219} use an ensemble of neural network models and \citet{ritter2018a} make use of the Laplace approximation to a trained network to obtain uncertainty estimates. In this work, we follow this train of thought and along with these papers empirically show that the methods that model weight dependencies improve prediction performance in terms of test likelihoods. 


An important application of the learned uncertainty estimates is model selection via model pruning. The concept of pruning, sparsifying, or compressing a network makes sense for deep neural network models as neural network models are often over-parameterized. 
%
The task of compressing a network has received much attention due to the immense demand on the efficient deployment of deep models on mobile devices.
%
For instance, \citet{louizos2017bayesian} consider group Horseshoe priors and log-Normal posteriors for pruning out neurons, and \citet{pmlr-v80-ghosh18a} improve the result of \cite{louizos2017bayesian} by adopting regularized Horseshoe priors and more structured posteriors.
\citet{neklyudov2017structured} take a slightly different approach which is pruning units via truncated log-Normal priors over unit scales. All of these methods provide significantly condensed networks after pruning out the neurons that are less contributing than the rest. 



\subsection*{Our contributions}

In this paper, we propose a new variational family with the aim of not only tackling the modelling side of BNNs in terms of capturing correlations among weight parameters but also addressing the issue of sparsification of over-parameterized neural network models.  Our variational family has two components, where each component is built for taking care of each of these aspects. Since we can decompose a given weight vector into its magnitude (radius) and angle (direction), we decompose the variational posterior distribution into radial and directional densities. 
%
%
This combination of prior and posterior pairs have the following benefits. 

%


\paragraph{Directional density.} We use the von Mises-Fisher (vMF) distribution to capture correlations between the weight parameters. 
While the vMF distribution is a typical choice for directional data modelling \cite{banerjee2005clustering}, 
applying it in variational inference
poses a practical challenge due to difficulty in computing ratios of Bessel functions (not supported yet in most deep learning frameworks).
%
We propose an approximation to the polynomials of ratios of modified Bessel functions for numerically stable gradient estimation, which is scalable to high dimensions such as several hundreds and a few thousands.
This approximation can be applied to general variational inference problems using vMF distributions in high dimensional spheres.

\paragraph{Radial density.} We use the Half-Cauchy prior and the log-Normal posterior to provide a structured compression method for neural network parameters. 
The Half-Cauchy prior is chosen to provide a bias toward zero in the norm of weights. 
The log-Normal posterior is chosen to provide a closed-form KL divergence between prior and posterior in the computation of evidence lower bound. 

Equipped with these two components, our ultimate goal is to find a {\it{good}} model for a dataset, via finding the dependencies between the weights and reducing the number of parameters. So the goodness of a model will be measured by test likelihoods and the size of the resulting model (e.g., in terms of number of remaining parameters and FLOPs. See \secref{experiments} for details).

The rest of the paper is organized as follows. We start by describing essential quantities in variational Bayesian learning for neural network models in \secref{background}. In \secref{method}, we introduce our new variational family and address how we made the training under the new model numerically stable with high dimensional vMF. In \secref{relatedwork}, we contrast our method with other existing work. Finally in \secref{experiments} we empirically show that our method improves predictive ability over existing methods together with structured pruning.   


\section{Variational Bayesian learning for neural networks}\label{sec:background}

Consider a neural network consisting of fully-connected\footnote{For notational simplicity, we stick to a network with fully-connected layers. See \subsecref{l2_control} for details on the implementations for networks with convolutional layers.} layers with a collection of parameters $\mW = \{\mW^{(l)}\}_{l=1, \cdots, L}$, where the $l$-th layer's weight matrix is denoted by $\mW^{(l)} \in \R^{n_l \times n_{l-1}}$. Here, $n_l$ is the number of output neurons and $n_{l-1}$ is the number of input neurons. 

In variational Bayesian neural networks, in an attempt to capture distributional behaviors of the weights, we often assume a tractable parametric family for the prior distribution $p_\vtheta(\mW)$ and the approximate posterior  $q_\vphi(\mW)$, where the parameters for each distribution are denoted by $\vtheta$ and $\vphi$, respectively.
Given a dataset $\Dat$, we maximize the variational (evidence) lower bound (ELBO) to the marginal data likelihood, given by,
\begin{align}
\label{eq:ELBO}
    \mathcal{L}(\Dat; \vphi, \vtheta) &= \Em_{q_\vphi(\mW)}[\log p(\Dat|\mW)] \nonumber \\
    & \qquad - D_{KL}(q_\vphi(\mW)|| p_\vtheta(\mW))
\end{align} in order to choose the parameters of prior and posterior distributions. The first term in \eqref{ELBO} is the expected log-likelihood with respect to the variational posterior. The second term in \eqref{ELBO} is the KL divergence between the variational posterior and the prior distribution.


We would like to point out two important conditions for the lower-bound optimization to be successful in practice. First, under neural network models, the expected log-likelihod term in \eqref{ELBO} does not have a close form and is typically estimated via Monte Carlo (MC) sampling \cite{hoffman2013stochastic, kingma2013auto}. The MC estimator, however, exhibits high variance in the gradients of expected log-likelihood. To alleviate this issue, many solutions have been proposed~\cite{kingma2013auto, maddison2016concrete, jang2016categorical, figurnov2018implicit, tucker2017rebar}.
The most known solution among them is {\it{reparametrization trick,}} which detaches the parameter part of  $q_{\vphi}(\mW)$ from the random source of $q_{\vphi}(\mW)$ so that the MC-estimate of the gradient of the expected log-likelihood can be computed independently of the randomness of the distribution. For instance, for $\mW \sim q_\vphi(\mW)$, we reparameterize the distribution $q(\mW)$ as a deterministic function $z(\epsilon, \vphi)$ where $\epsilon$ is a parameter-free random sampler $\epsilon \sim s(\epsilon)$ (independent of $\vphi$), and hence the resulting gradient is independent of the random source. 
Using this reparametrization, we can effectively reduce the variance of the MC-estimate of the gradient of log-likelihood.

Secondly, for arbitrary priors $p(\mW)$ and posteriors $q(\mW)$, similar MC-estimates using the reparametrization trick can be found \cite{blundell2015weight}. However, in practice, many models propose closed-form KL-divergence to further reduce the variance of the gradient estimates (e.g., \cite{louizos2017bayesian}). 
In summary, being able to use the reparameterization trick and having a closed-form KL divergence are important conditions for variational learning to be effective in BNNs. We will revisit this point in \secref{method}. 

From a modelling perspective, specifying the functional forms of the prior and posterior distributions
is an essential step to perform variational BNNs.
%
One of the most commonly used variational family is fully factorized distribution referred to as the {\it{mean-field}} variational family  \cite{graves2011practical, blundell2015weight, kingma2015variational, neklyudov2017structured, ullrich2017soft, molchanov2017variational}
   $ q(\mW) = \prod_{l=1}^L \prod_{w \in \mW^{(l)}} q(w)$,
where the posterior distribution is described by a product of distributions of each individual entry in $\mW$.
%
This choice is due to computational tractability. Consider the Gaussian mean-field variational family, $q(w_{i,j}^{(l)}|\vphi_{i,j}) = \Nrm(\mu_{i,j}, \sigma^2_{i,j})$, where the {\it{variational}} parameters are $\vphi_{i,j} = \{\mu_{i,j}, \sigma^2_{i,j}\}$. 
This formulation allows separate gradient updates for each of the variational parameters 
as there is no dependencies between them. 
However, this variational family ignores any statistical correlations between the weight parameters, which is the issue we address in this paper.


Next we describe our new variational family which we developed with all these computational aspects taken under consideration, namely, the reparameterizability of expected log-likelihood term, a closed-form KL divergence term, and separate gradient updates for each of the variational parameters.

\section{Method}\label{sec:method}

\subsection{Radial and Directional Posterior}\label{subsec:new_va_fam}

We propose a new variational family, which is an instance of {\it{structured}} mean-field approximation where each row (and/or column) of $\mW^{(l)}$ is factorized, 
\begin{equation}
    q(\mW) = \prod_{l=1}^L \prod_{r=1}^{n_l} q_r^{(l)}(\vw_{r}^{(l)}),
\end{equation} where $\vw_{r}^{(l)} \in \mathbb{R}^{n_{l-1}}$ is the $r$-th row of the weight matrix at $l$-th layer.
For the sake of simplicity, we consider row-wise factorization here. But we discuss both row and column-wise factorizations later in \subsecref{l2_control}.

We decompose each factor $q_r^{(l)}$ into  a density on the $L_2$-norm of $\vw_{r}^{(l)}$ and another density on the direction of the normalized vector $\vw_{r}^{(l)}/ \Vert \vw_{r}^{(l)} \Vert_2$, given as
\begin{equation}
    q_r^{(l)}(\vw_{r}^{(l)}) = q_{r, rad}^{(l)}(\Vert \vw_{r}^{(l)} \Vert_2) \cdot q_{r, dir}^{(l)} \left( \frac{\vw_{r}^{(l)}}{\Vert \vw_{r}^{(l)} \Vert_2} \right)
\end{equation}
where $q_{r, rad}^{(l)}$ is the \textit{radial density} and $q_{r, dir}^{(l)}$ is the \textit{directional density}. We now introduce each of these densities in detail. 



\begin{figure}[t]
\vspace{-0.4em}
\includegraphics[width=0.5\textwidth]{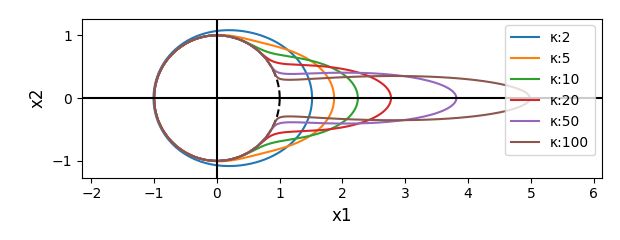}
\vspace{-3em}
\caption{Von Mises (Fisher) distribution on $\mathcal{S}^1$ (i.e., a probability distribution on the directional component of a 2-dimensional vector) for various concentration parameters $\kappa$ with a fixed location parameter $\vmu = (1, 0)$. With a low level of concentration (blue trace), the probability mass is widely spread from the center location. As we increase the level of concentration from $2$ to $100$ (from blue to brown traces), the probability density is getting highly concentrated around the center location.}
\label{fig:vmf_S1}
\vspace{-1em}
\end{figure}

\vspace{-1em}
\paragraph{Directional density}
We take the prior $p_{r, dir}^{(l)}$  and the posterior $q_{r, dir}^{(l)}$ distributions to be the {\it{von Mises-Fisher}} (vMF) distribution \cite{mardia2009directional}, which is a probability density on a unit (hyper)sphere, and its density is given by 
\begin{align}
  q_{r, dir}^{(l)} \left(\frac{\vw_{r}^{(l)}}{\Vert \vw_{r}^{(l)} \Vert_2}\right) &= vMF \left(\frac{\vw_{r}^{(l)}}{\Vert \vw_{r}^{(l)} \Vert_2} \Big\vert \vmu_r^{(l)}, \kappa_r^{(l)} \right),
  \end{align} 
where the probability density function of a vMF distributed $d$-dimensional unit vector $\vx$ is given by 
$vMF(\vx \vert \vmu, \kappa) = C_d(\kappa) \exp{(\kappa \vmu^T \vx)}$, where $
    C_d(\kappa) = \frac{\kappa^{d/2-1}(\kappa)}{(2\pi)^{d/2}\mathcal{I}_{d/2-1}(\kappa)}.$
The location parameter $\vmu$ is also a $d$-dimensional unit vector, $\kappa$ is the concentration parameter, and $I_{d/2-1}$ is the modified Bessel function of the first kind at order $d/2-1$. The vMF distribution is intuitively understood as multivariate Gaussian distribution with a diagonal covariance matrix on unit (hyper)sphere. To help the readers gain intuition how the vMF distribution behaves, we illustrate a 2D vMF density with different values of concentration in \figref{vmf_S1}.

In our prior and posterior distributions, we assume that the concentration parameter is shared across all the rows in each layer, by assigning a single concentration parameter, $\kappa_r^{(l)} = \kappa^{(l)}$ for all $r = \{1, \cdots, n_l\}$, while the mean vector parameters are separately assigned for each row.  This way we can reduce the number of variational and prior parameters significantly. The resulting prior $p_{r, dir}^{(l)}$ and posterior $q_{r, dir}^{(l)}$ distributions have the following forms:
\begin{align}
p_{r, dir}^{(l)} \left(\frac{\vw_{r}^{(l)}}{\Vert \vw_{r}^{(l)} \Vert_2}\right) &= vMF \left( \vmu_{p,r}^{(l)}, \kappa_p^{(l)} \right), \\
q_{r, dir}^{(l)} \left(\frac{\vw_{r}^{(l)}}{\Vert \vw_{r}^{(l)} \Vert_2}\right) &= vMF \left(\vmu_{q, r}^{(l)}, \kappa_{q}^{(l)} \right),
\end{align} where the prior mean parameters
and the concentration parameter are denoted by $\vmu_{p,r}^{(l)}$ and $\kappa_{p}^{(l)}$, respectively, and those in the posterior distribution are denoted by $\vmu_{q, r}^{(l)}$, and $\kappa_{q}^{(l)}$. 

Explicitly modelling the directional component using vMF allows us to capture the correlation within the weight parameters of each row. Interestingly, having the same concentration parameter across all the rows within each layer induces a particular way of correlations in the weight parameters within the same layer. For instance, if the mean parameters of each row's weights are somewhat close to each other, having the same concentration level, possibly a high concentration level (which we expect if the posterior confidence is high) makes the row-wise directional densities {\it{more correlated}} with each other. On the other hand, if the mean parameters of each row's weights are all far from each other, having the same high concentration level makes the row-wise directional densities {\it{less correlated}} with each other. Surely, there are other ways to model correlations across rows. However, from our experience, this particular way of parameterizing the variational parameters allows us to capture the correlation across rows effecively (as shown in \secref{experiments}) without assigning concentration parameters to each of the rows and layers. 






\paragraph{Radial density}

\begin{figure}[t]
\centering
\includegraphics[width=0.3\textwidth]{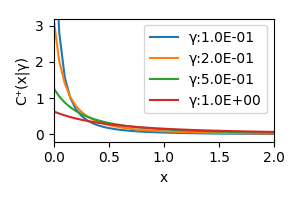}
\vspace{-2em}
\caption{The probability density function of Half-Cauchy distribution $\mathcal{C}^+(\cdot \vert \gamma)$ for a different value of scale parameter $\gamma$ (inset). The smaller $\gamma$ gets, the larger the probability mass near 0 gets, yielding sparsity.}\label{fig:halfcauchy}
\end{figure}

While we could adopt any probability distribution with a non-negative support for the radial density, we focus on distributions that can promote sparsity in the resulting posterior. Specifically, inspired by the group horseshoe prior proposed by~\citet{louizos2017bayesian},
we take a product of two Half-Cauchy distributions to be our prior in order to induce sparsity in the {\it{norms}} of the weights. First, we write down the norm of each row given a layer as a product of two independent  {\it{half-Cauchy}} random variables,
\begin{gather}
    \Vert \vw_{r}^{(l)} \Vert_2 = s^{(l)} \bar{z}_{r}^{(l)}; \nonumber \\
    s^{(l)} \sim \mathcal{C}^+(\gamma) \quad \bar{z}_{r}^{(l)} \sim \mathcal{C}^+(1),
\end{gather} and  the prior 
is given by 
\begin{align}
p_{r, rad}^{(l)}(\Vert \vw_{r}^{(l)} \Vert_2) & =  \mathcal{C}^+(s^{(l)} \vert \gamma) \cdot \mathcal{C}^+(\bar{z}_{r}^{(l)} \vert 1),
\end{align} 
where the probability density function for a half-Cauchy distributed random variable $x$ is given by 
 $   \mathcal{C}^+(x \vert \gamma) = \frac{2}{\pi \gamma (1 + (x/\gamma)^2)},$
with a scale parameter $\gamma >0$. As illustrated in \figref{halfcauchy}, the smaller the scale parameter gets, the larger the probability mass concentrates around zero.
At this point, it might not be immediately clear why we chose to use two Half-Cauchy distributions as a prior rather than one. Our explanation is as follows. 

What we ultimately hope to control is the level of sparsity in the weights drawn from the resulting posterior distribution. We allow the posterior to have two different levels of sparsity, namely, \textit{local}  (row-wise) sparsity and \textit{global} (layer-wise) sparsity.
The reason we write the norm as a product of two terms $\Vert \vw_{r}^{(l)} \Vert_2 = s^{(l)} \bar{z}_{r}^{(l)}$ is that each of these terms affects the local sparsity via $\bar{z}_{r}^{(l)}$ and global sparsity via  $s^{(l)}$ in the posterior distribution, respectively. 
Even when all radii are small, the largest one among them has significant influence in model performance.
Thus, we can use the relative strength of radius densities to prune out.

Before describing our radial posterior density, we need to elaborate on the fact that each of half-Cauchy distributions can be further factorized. As used in \citet{louizos2017bayesian}, the half-Cauchy distribution can be written as a product of an inverse-Gamma density and a Gamma density due to the fact that
the square of half-Cauchy $\mathcal{C}^+$ is equal in distribution to a product of Gamma and inverse-Gamma~\cite{neville2014mean}. As a result, we can rewrite the two factors $\bar{z}_{r}^{(l)}$ and $s^{(l)}$ for describing $\Vert \vw_{r}^{(l)} \Vert_2$ in the prior distribution as 
\begin{align}
\sqrt{\bar{z}_{r}^{(l)}} &= \alpha_{r}^{(l)} \beta_{r}^{(l)}, \mbox{where } \alpha_{r}^{(l)} \sim  \mathcal{G}(\tfrac{1}{2}, 1), \beta_{r}^{(l)} \sim \mathcal{IG}(\tfrac{1}{2}, 1)  , \nonumber \\
    \sqrt{s^{(l)}} &= s_a^{(l)} s_{b}^{(l)}, \mbox{where } s_{a}^{(l)} \sim  \mathcal{G}(\tfrac{1}{2}, \gamma^2), s_{b}^{(l)} \sim \mathcal{IG}(\tfrac{1}{2}, 1), \nonumber
\end{align} 
where we denote the Gamma distribution by $\mathcal{G}$ and the inverse-Gamma distribution by $\mathcal{IG}$. 
In addition, Gamma (and inverse-Gamma) random variables have another nice property, that is, one can obtain a closed-form expression of KL divergence between a Gamma (and inverse-Gamma) distribution and a log-normal distribution. Using this fact, we take the log-normal distribution to be our posterior such that each of these terms can control the local sparsity and global sparsity, given as 
\begin{align}\label{eq:radius_posterior}
    &q_{r,rad}^{(l)} (\Vert \vw_{r}^{(l)} \Vert) \nonumber \\
    &= \mathcal{LN}(\alpha_{r}^{(l)}| \mu_r^{(l)}, \sigma_r^{2\,(l)}) \mathcal{LN}( \beta_{r}^{(l)}| \widehat{\mu}_r^{(l)}, \widehat{\sigma}_r^{2\,(l)}) \nonumber \\
    & \qquad \mathcal{LN}(s_a^{(l)} | \mu^{(l)}, \sigma^{2\,(l)}) \mathcal{LN}(s_b^{(l)} | \widehat{\mu}^{(l)}, \widehat{\sigma}^{2\, (l)}), 
\end{align} 
where $\mu_r^{(l)}$, $\sigma_r^{2\,(l)}$ and $\widehat{\mu}_r^{(l)}$, $\widehat{\sigma}_r^{2\,(l)}$ are the variational parameters approximating Gamma density and inverse-Gamma density of $\mathcal{C}^+(\bar{z}_{r}^{(l)} \vert 1)$, respectively. 
Similarly, $\mu^{(l)}$, $\sigma^{2\,(l)}$ and $\bar{\mu}^{(l)}$, $\widehat{\sigma}^{2\,(l)}$ are variational parameters for the prior $\mathcal{C}^+(s^{(l)} \vert \gamma)$ the same way.



\begin{figure}[t]
    \centering\includegraphics[width=0.4\textwidth]{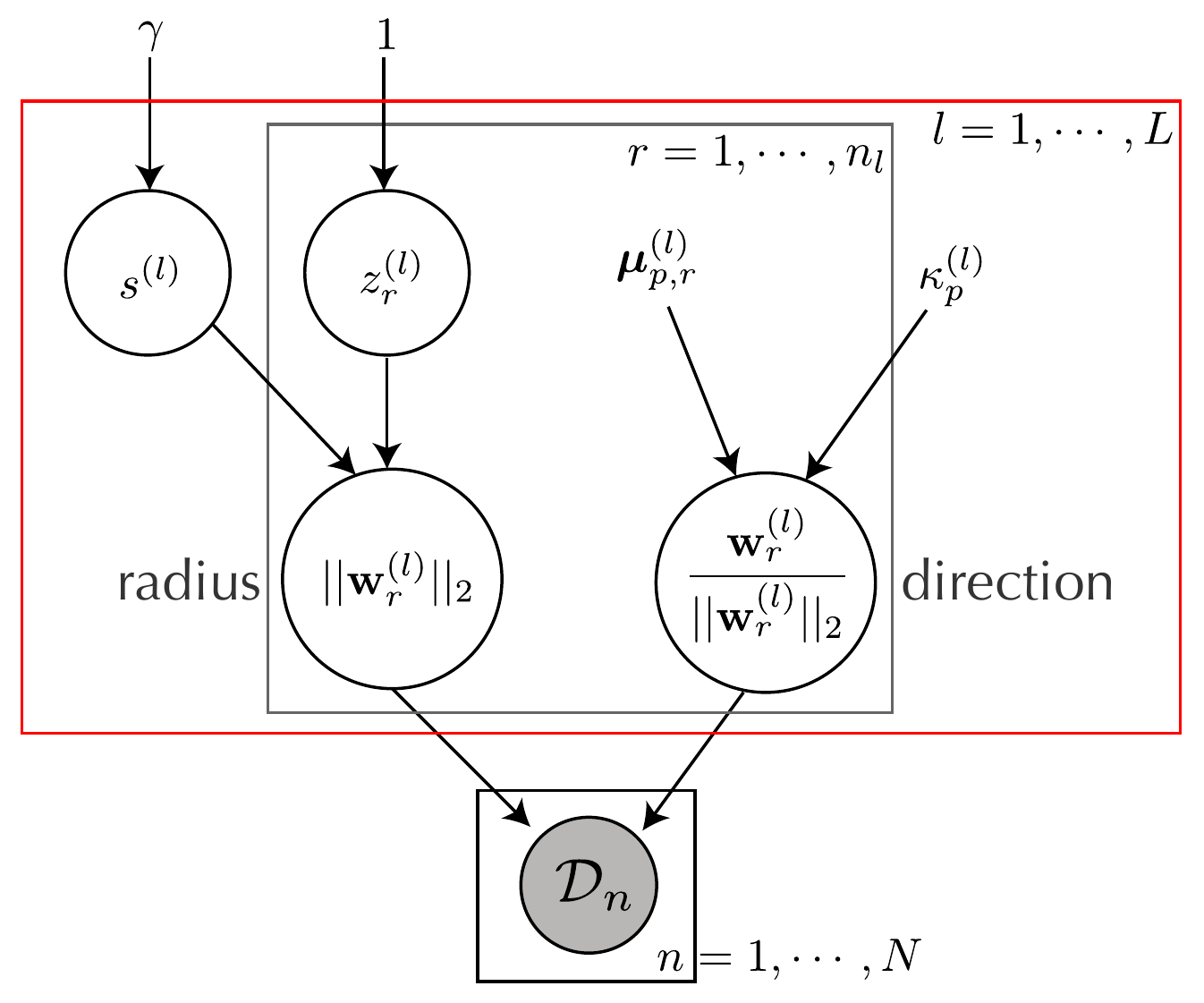}
\caption{Graphical representation of our generative model. The black rectangle describes row parameters and red rectangle layer parameters. The radius part is controlled by two half-Cauchy random variables $s^{(l)}$ and $z_r^{(l)}$, while the directional part is controlled by a vMF distribution governed by the prior mean vector and a concentration parameter, $\vmu_r^{(l)}$ and $\kappa_p^{(l)}$, respectively. The two parts taken together govern the data probability (bottom). }
\label{fig:graphical_model}
\vspace{-1.5em}
\end{figure}

\begin{figure}[t]
\centering\includegraphics[width=0.45\textwidth]{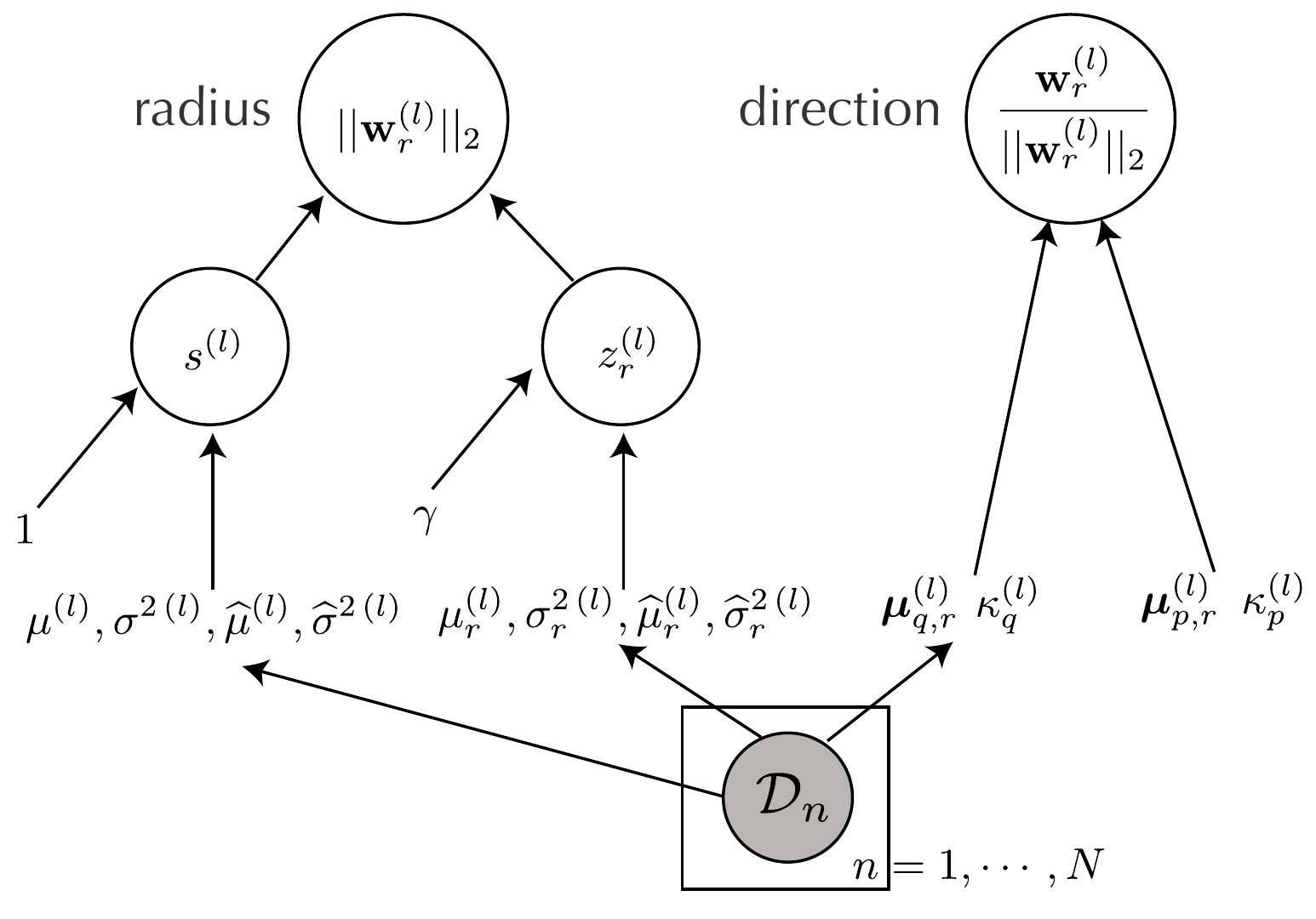}
\caption{Variational inference with radial and directional posterior (RDP). Given a dataset $\Dat$ (bottom), by maximizing the variational lower bound, we estimate the variational parameters (middle), which, together with the prior terms, determine the posterior distribution over the radial and directional components (top). }
\label{fig:posterior_inference}
\vspace{-1em}
\end{figure}

\paragraph{Radial and directional posterior (RDP)}
In summary, our variational family $q(\mW)$ is given as
\begin{align}\label{eq:radial_mean_field}
   q(\mW) =  \prod_{l=1}^L &\mathcal{LN}(s_a^{(l)} | \mu^{(l)}, \sigma^{2\,(l)}) \mathcal{LN}(s_b^{(l)} | \widehat{\mu}^{(l)}, \widehat{\sigma}^{2\,(l)}) \nonumber \\
   & \prod_r^{n_l} \mathcal{LN}(\alpha_{r}^{(l)}| \mu_r^{(l)}, \sigma_r^{2\,(l)}) \mathcal{LN}( \beta_{r}^{(l)}| \widehat{\mu}_r^{(l)}, \widehat{\sigma}_r^{2\,(l)}) \nonumber \\
   & \quad \cdot vMF \left(\frac{\vw_r^{(l)}}{\Vert \vw_r^{(l)} \Vert_2}; \vmu_{q,r}^{(l)}, \kappa_{q}^{(l)} \right).
\end{align}  We refer to this collection of densities as the {\it{radial and directional posterior (RDP)}}.
Our prior distribution $p(\mW)$ is also factored into radius and direction as in the proposed variational family, given as
\begin{align}
     p(\mW) & = \prod_{l=1}^L \mathcal{C}^+(s^{(l)} ; \gamma)\prod_r^{n_l} \mathcal{C}^+ \left(\bar{z}_{r}^{(l)}; 1 \right) \nonumber \\
     & \qquad \cdot vMF \left( \frac{\vw_r^{(l)}}{\Vert \vw_r^{(l)} \Vert_2}; \vmu_{p,r}^{(l)}, \kappa_p^{(l)} \right).
\end{align} 
%
We denote the prior parameters collectively by $\vtheta = \{\gamma,  \vmu_{p,r}^{(l)}, \kappa_p^{(l)} \}$ for all layers $l=\{1,\cdots,L\}$, and the variational parameters by $\vphi = \{ \mu^{(l)}, \sigma^{2\,(l)}, \widehat{\mu}^{(l)}, \widehat{\sigma}^{2\,(l)}, \mu_r^{(l)}, \sigma_r^{2\,(l)}, \widehat{\mu}_r^{(l)}, \widehat{\sigma}_r^{2\,(l)}, \vmu_{q,r}^{(l)}, \kappa_{q}^{(l)}\}$ for all layers $l=\{1,\cdots,L\}$ and all rows $r = \{1, \cdots, n_l \}$.
A graphical representation of our model is given in \figref{graphical_model} for the generative process and in \figref{posterior_inference} for the inference process.

\subsection{Optimizing evidence lower bound for RDP}

Recall that as far as our objective function \eqref{ELBO}  is concerned,
two conditions need to be met for the gradients of this objective function to well behave. The first condition (about MC estimates of the expected log-likelihood term) is whether our posterior is reparameterizable. In fact, we  can represent our choice of posterior by
a differentiable function $h(\epsilon, \vphi)$, where the variational parameters $\vphi$ are separated from the random source, $\epsilon \sim s(\epsilon)$. See \suppsecref{repara_vMF} for details.


%

The second condition is whether the KL term is closed-form, which is the case as we choose the prior and posterior pair considering this condition. The KL term is given by
\begin{align}
    &D_{KL}(q_{\vphi}(\mW) \Vert p_{\vtheta}(\mW)) = \nonumber \\
    &\sum_l \sum_r^{n_l} D_{KL}(vMF (\vmu_{q,r}^{(l)}, \kappa_{q,r}^{(l)} ) \Vert vMF (\vmu_{p,r}^{(l)}, \kappa_{p}^{(l)} )  ) \nonumber \\
    & \quad \qquad + D_{KL}(q_{r, rad}^{(l)}(\Vert \vw_{r}^{(l)} \Vert_2) \Vert p_{r, rad}^{(l)}(\Vert \vw_{r}^{(l)} \Vert_2)),
\end{align} and the closed-form expressions of each of the terms are given in Appendix. 
Although the KL term between the vMF prior and posterior is elegantly written in closed-form, 
\begin{align}
    &D_{KL}(vMF(\mu_q, \kappa_q) \Vert vMF(\mu_p, \kappa_p)) = \nonumber \\
    &(\kappa_q - \kappa_p\mu_p^T\mu_q) \frac{I_{d/2}(\kappa_q)}{I_{d/2-1}(\kappa_q)}  + \log(C_d(\kappa_q)) - \log(C_d(\kappa_p)) \nonumber
\end{align} the gradient expressions (see \suppsecref{kl_divergence}) with respect to the variational parameters require computing the ratio $\frac{I_{d/2}(\kappa_q)}{I_{d/2-1}(\kappa_q)}$ \cite{davidson2018hyperspherical}, which is numerically unstable. This is due to the fact that the modified Bessel function of the first kind (Bessel function) decays rapidly, so the computation of ratios of Bessel functions causes numerical errors when it tries to compute $\frac{0}{0}$ (See \suppsecref{instable_bessel} for detailed explanations). This gets worse with higher dimensions, and occurs even for moderate dimensions such as 50 to 100. Subsequently, we show how to address this issue of numerical instability. 



\vspace{-1em}
\paragraph{Overcoming numerical issues when using vMF distributions}


Rather than numerically computing the ratio of Bessel functions, we resort to the following Theorem,
\begin{theorem}[Theorem 5 in \cite{ruiz2016new}]\label{thm:bessel_ratio_bound}
    \begin{equation}
        B_2(\nu, z) < \frac{I_{\nu}(z)}{I_{\nu-1}(z)} < B_0(\nu, z), \quad \text{when} \quad \nu \ge 1/2
    \end{equation}
    \text{where\vspace{-0.6em}}
    {\small\begin{align}
        &B_{\alpha}(\nu, z) = \frac{z}{\delta_{\alpha}(\nu, z) + \sqrt{\delta_{\alpha}(\nu, z)^2 + z^2}} \nonumber \\
        &\delta_{\alpha}(\nu, z) = (\nu - 1/2) + \frac{\lambda}{2\sqrt{\lambda^2+z^2}}, \qquad \lambda=\nu+(\alpha-1)/2 \nonumber
    \end{align}}\normalsize
\end{theorem} 
\vspace{-1em}
where $\nu$ denotes the dimension and $z$ denotes the concentration parameter.

\begin{figure}[t]
\vspace{-0.6em}
\centering{\includegraphics[width=0.4\textwidth]{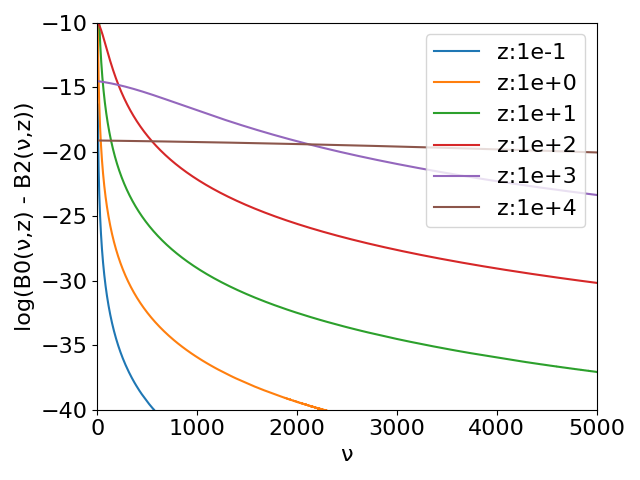}
\vspace{-1em}
\caption{Tightness of bounds in Thoerem \ref{thm:bessel_ratio_bound}. For large dimensions $\nu$, the bounds become very tight, resulting in almost no gap between the lower and upper bounds. Note that the largest difference between two bounds is approximately $e^{-10}$.}}
\label{fig:tight_bound}
\vspace{-1.6em}
\end{figure}

Our observation is that
the gap between the upper and lower bounds of the ratio becomes tighter as the dimension grows as shown in Fig.\ref{fig:tight_bound}.
Even in low dimensions, the gap is less than $e^{-10}$ for various concentration parameter  values ($z$).
Using this fact, we simply approximate the ratio by the average over the lower and upper bounds, 
$\frac{I_{\nu}(z)}{I_{\nu-1}(z)} \approx (B_2(\nu, z) + B_0(\nu, z))/2$.

Empirically we find that this simple approximation allows us to obtain numerically stable gradients on dimensions of several thousands.
Furthermore, this approximation saves us from directly computing modified Bessel function.
Since the modified Bessel function of the first kind of high order is not supported yet in most deep learning frameworks, using this approximation, variational inference with high dimensional vMF distributions can enjoy GPU acceleration without extra efforts on CUDA implementations of Bessel functions.

\vspace{-1em}
\subsection{Structured compression using the radial density }\label{subsec:l2_control}
\begin{figure}[t]
\vspace{-1.0em}
\centering\includegraphics[width=0.4\textwidth]{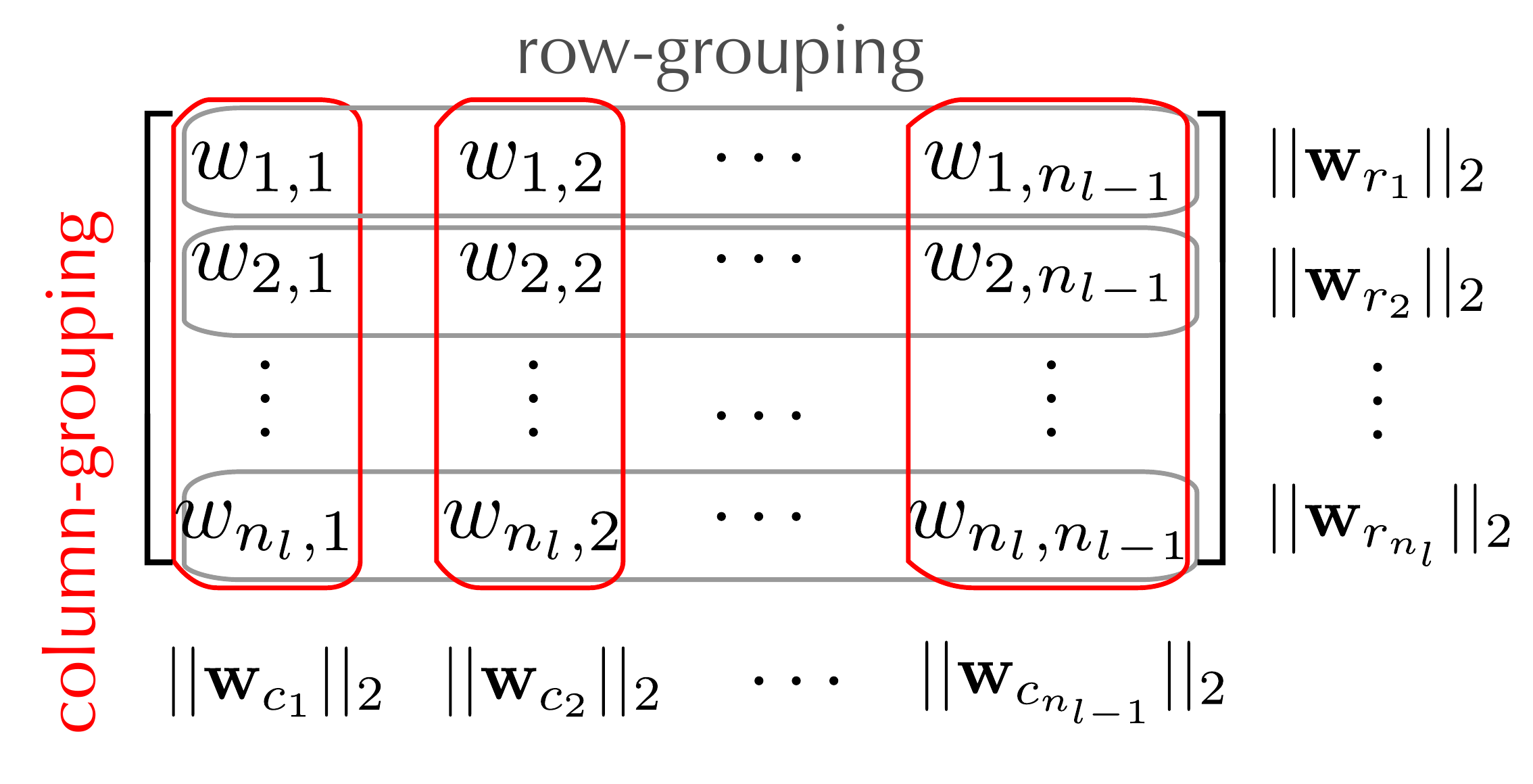}
\vspace{-1em}
\caption{Column grouping in red to prune out input neurons, depending on L2-norm of each column vector. Row grouping in grey to prune out output neurons. Double grouping in red and grey to prune out both input and output neurons. }
\label{fig:Grouping}
\vspace{-1.4em}
\end{figure}

\vspace{-0.4em}
Subsequently to presenting the RDP family and overcoming the optimization issues, we show the utility of the radius component in search for a more optimal model in form of model compression.  Thus, using the \textit{radial} density given in \eqref{radial_mean_field}, we can prune out output neurons on each layer depending on the contribution of each neuron measured by the learned posterior distribution. We call this scheme {\it{row-grouping}} as depicted in \figref{Grouping}.
One can employ different types of grouping. For instance, if pruning out input neurons is of interest, the radial and directional posteriors can be formulated in terms of {\it{column-grouping}}, given by 
\vspace{-0.6em}
\begin{align}
 q(\mW) &= \prod_{l=1}^L \prod_{c=1}^{n_{l-1}} q_c^{(l)}(\vw_{c}^{(l)}), \\
        q_c^{(l)}(\vw_{c}^{(l)}) &= q_{c, rad}^{(c)}(\Vert \vw_{c}^{(l)} \Vert_2) \cdot q_{c, dir}^{(l)} \left( \frac{\vw_{c}^{(l)}}{\Vert \vw_{c}^{(l)} \Vert_2} \right).
\end{align} 
One can also prune out both input and output neurons simultaneously,  in which case the radial and directional posteriors can be formulated in terms of {\it{double-grouping}}, given by 
\vspace{-0.4em}
\begin{align}
 q(\mW) &= \prod_{l=1}^L \prod_{r=1}^{n_{l}} \prod_{c=1}^{n_{l-1}} q_r^{(l)}(\vw_{r}^{(l)}) q_c^{(l)}(\vw_{c}^{(l)}), \\
        q_r^{(r)}(\vw_{r}^{(l)}) &= q_{c, rad}^{(r)}(\Vert \vw_{r}^{(l)} \Vert_2) \cdot q_{r, dir}^{(l)} \left( \frac{\vw_{r}^{(l)}}{\Vert \vw_{r}^{(l)} \Vert_2} \right), \\
    q_r^{(c)}(\vw_{c}^{(l)}) &= q_{c, rad}^{(c)}(\Vert \vw_{c}^{(l)} \Vert_2) \cdot q_{c, dir}^{(l)} \left( \frac{\vw_{c}^{(l)}}{\Vert \vw_{c}^{(l)} \Vert_2} \right).
\end{align} The prior distribution also needs to be modified according to which grouping the posterior distribution takes. Different grouping schemes are illustrated in \figref{Grouping}.

Note that for a neural network with convolutional layers, $\mW^{(l)}$ is a convolutional filters of 4D tensor of $n_l \times n_{l-1} \times k_w \times k_h$ where $n_{l-1}$ is the number of input channels, $n_l$ is the number of output channels and $k_w$ and $k_h$ are kernel width and height.
Then row-wise grouping of the filter $\mW^{(l)}$ is row-wise grouping of 2D flattened matrix of dimension $n_l \times (n_{l-1} \cdot k_w \cdot k_h)$ which is grouped by an output channel. 
Column-wise grouping is to group by input channels, grouping of the flattened matrix of dimension $(n_l \cdot k_w \cdot k_h) \times n_{l-1}$.

Following \cite{louizos2017bayesian}, we use the (log of) posterior mode (which is the mean parameter minus the variance parameter of the log-normal posterior distribution) as a cut-off threshold to determine which output neuron needs to remain or be pruned out from the model. 
Recall that there are four log-normal distributions to approximate the posterior over the radius in \eqref{radius_posterior}. 
Since a product of two log-normals is a log-normal with summed-up parameters from the two, we use the mode of this combined log-normal for pruning.
Note that for pruning rows (or columns or both) only the two 'local' log-normal distributions in \eqref{radius_posterior} matter as the remainders are 'global' ones (i.e., the global ones scale up and down the local ones in the exact same way across rows/columns). 
\vspace{-0.6em}
\section{Related Work}\label{sec:relatedwork}
\vspace{-0.4em}

As mentioned in \secref{intro} and \secref{background}, many existing papers on variational BNNs assume the mean-field posterior distribution on the network weights.
Recently, \citet{louizos2016structured, sun2017learning, zhang2017noisy, sun2018functional} proposed to take dependency between weights into account, e.g., \citet{louizos2016structured} uses a matrix normal posterior with diagonal covariance matrices.
However, unlike \cite{louizos2016structured, sun2017learning, zhang2017noisy, sun2018functional}, we capture dependency between weights through a directional component by restricting it to be on a unit sphere, which impose dependencies on weights (within a row and/or a column) directly. In addition, by coupling the concentration parameters per layer, we capture correlations among weights across rows, columns, and/or both.  

\vspace{-0.2em}
The reparametrization trick is an essential tool in variational learning of BNNs.
When it comes to von Mises-Fisher distribution, 
only the vMF reparametrization trick proposed by \citet{davidson2018hyperspherical} that implements rejection sampling \cite{naesseth2016reparameterization} is practically applicable, due to the complexity of the sampling procedure for vMF. 
In our case, using the vMF posterior introduces a new challenge as scalability to high dimension is required. We improve the above approaches with a simple approximation method which makes the vMF reparametrization trick scalable up to several thousands of dimensions. 
Besides, the approximation yields MC gradient estimates that are corrected to be unbiased for more efficient and stable training (see \suppsecref{repara_vMF} for details).







\vspace{-0.2em}
Research on neural network compression started with  non-Bayesian approaches (e.g, \cite{hassibi1993second}) and mostly focused on non-structured pruning (e.g., \cite{han2015deep}).
%
%
%
Recently, hardware-oriented considerations have turned the research towards structured pruning for more practical speed-ups (e.g., \cite{srinivas2015data,li2016pruning, wen2016learning,Lebedev_2016, zhou2016less}).
%
Subsequently by taking into account the network weights' uncertainties, Bayesian methods 
have achieved an impressive compression rate. For example, 
\citet{molchanov2017variational} proposed the mean-field Gaussian posterior along with sparsity inducing prior on scale parameters; 
\citet{ullrich2017soft} proposed a Gaussian mixture prior; and 
\citet{louizos2017bayesian} proposed structurely grouping weights through a group Horseshoe prior. 
%
%
In contrast, our RDP utilizes the probability density over the magnitudes of weights and prune out neurons based on captured uncertainties over the magnitudes.  


\vspace{-0.6em}
\section{Experiments}\label{sec:experiments}
\vspace{-0.4em}
Here, we provide empirical evidences supporting Radial and Directional Posteriors(RDP)'s strengths.
In the prediction task on UCI datasets, we compare RDP with mean-field based BNNs~\cite{graves2011practical, blundell2015weight, hernandez2015probabilistic} and a BNN designed to capture dependency~\cite{louizos2016structured} in order to show RDP's capability to explain dependency between weights.
In the compression task on MNIST dataset with LeNet arcthitecture, in order to check whether effective RDP's structure accommodates compression tasks, we compare RDP with various compression methods in terms of the amount of pruning and FLOPs.
In all experiment, we use Adam optimizer~\cite{kingma2014adam} with Pytorch default setting. 
In both tasks, double grouping was used. 
The code is available at \href{https://github.com/ChangYong-Oh/RadiusDirectionPosteriors.git}{https://github.com/ChangYong-Oh/RadiusDirectionPosteriors.git}

\vspace{-0.4em}
\subsection{Regression using UCI data}

\setlength{\tabcolsep}{4pt}
\begin{table}[t]
\vspace{-0.6em}
\begin{center}
\begin{footnotesize}
\begin{tabular}{ccccc}
\toprule
 \textbf{VI} & \textbf{PBP} & \textbf{Dropout} & \textbf{VMG} & \textbf{RDP} \\
\midrule
 -2.90$\pm$.07 & -2.57$\pm$.09 & -2.46$\pm$.25 & \textbf{-2.46$\pm$.09} & -2.60$\pm$.03\\

 -3.33$\pm$.02 & -3.16$\pm$.02 & -3.04$\pm$.09 & -3.01$\pm$.03 & \textbf{-2.61$\pm$.02}\\

 -2.39$\pm$.03 & -2.04$\pm$.02 & -1.99$\pm$.09 & \textbf{-1.06$\pm$.03} & -1.18$\pm$.03\\

 ~0.90$\pm$.01 & ~0.90$\pm$.01 & ~0.95$\pm$.03 & ~1.10$\pm$.01 & \textbf{~2.17$\pm$.00}\\

 ~3.37$\pm$.12 & ~3.73$\pm$.01 & ~3.80$\pm$.05 & ~2.46$\pm$.00 & ~3.05$\pm$.01\\

 -2.89$\pm$.01 & -2.84$\pm$.01 & -2.80$\pm$.15 & -2.82$\pm$.01 & \textbf{-0.14$\pm$.01}\\

 -2.99$\pm$.01 & -2.97$\pm$.00 & -2.89$\pm$.01 & -2.84$\pm$.00 & \textbf{~1.38$\pm$.01}\\

 -0.98$\pm$.01 & -0.97$\pm$.01 & -0.93$\pm$.06 & -0.95$\pm$.01 & \textbf{-0.45$\pm$.00}\\

 -3.43$\pm$.16 & -1.63$\pm$.02 & -1.55$\pm$.12 & \textbf{-1.30$\pm$.02} & -2.36$\pm$.04\\
\midrule
\end{tabular}
\end{footnotesize}
\vspace{-1em}
\caption{Average test log-likelihood on UCI regression tasks. Our method (RDP) achieves the better test likelihoods (five out of nine datasets) than other methods. The results in each row respectively come from tests on: Boston, Concrete, Energy, Kin8nm, Naval, Power plant, Protein, Wine, Yacht.} \label{tbl:exp_regression}
\end{center}
\vspace{-2em}
\end{table}

We compare the predictive performance on regression tasks UCI dataset tested in~\cite{gal2016dropout, louizos2016structured} following the experimental setting from~\cite{hernandez2015probabilistic}. 
We split the datasets so that $90\%$ is training data and $10\%$ is test data.\footnote{We followed the splitting rule given in \hyperlink{https://github.com/yaringal/DropoutUncertaintyExps}{https://github.com/yaringal/DropoutUncertaintyExps}}
On the prior on prediction noise, we use the same method as others detailed as follows. 
We use a Gamma prior for prediction precision $\tau$, $p(\tau)=\mathcal{G}(a_0=6, b_0=6)$ and posterior $q(\tau)=\mathcal{G}(a_1, b_1)$ for the precision of the Gaussian likelihood. We optimze $a_1, b_1$ along with all the other variational parameters.
The architecture we used is $n_{input} - 50 - 1$, so in this case for the second layer whose output dimension is one, we used fully factorized Gaussian and RDP with double grouping is only applied to the first layer.
Only with the change in the first hidden layer, we can see improvement over mean-field based BNNs, such as, Variational Inference (VI)~\cite{graves2011practical}, Probabilistic BackPropagation(PBP)~\cite{hernandez2015probabilistic}, Dropout~\cite{gal2016dropout}.
Compared to another dependency awaring posterior, Variational Matrix Gaussian (VMG)~\cite{louizos2016structured}, 5 out of 9 dataset, RDP shows better test log-likelihood(LL).

For more extensive comparison on this task with multi-layer neural network and deep Gaussian processes~\cite{damianou2013deep}, please refer to~\cite{bui2016deep}, in which you can observe the model with different RMSE and LL behaviors.

\subsection{Compression on MNIST Classification}

\begin{figure}[t]
\vspace{-0.4em}
\centering{\includegraphics[width=0.4\textwidth]{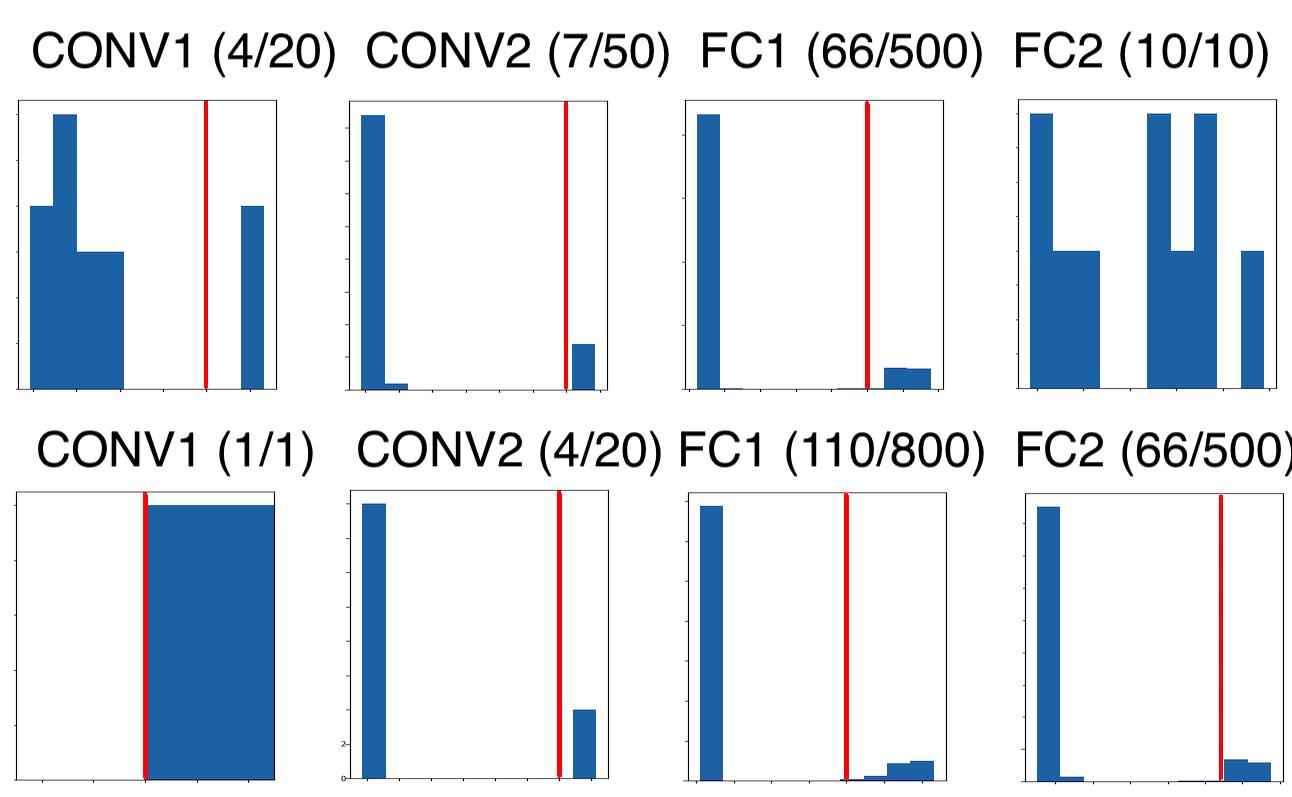}
\vspace{-0.6em}
\caption{The results of RDP compression. The horizontal axis is the log of mode and vertical axis is the number of rows (top) or columns (bottom) with that mode. The red dotted line is the pruning threshold. The numbers describe the ratio of unpruned weights/filters.}}
\label{fig:lenet5_compression}
\vspace{-1.5em}
\end{figure}

We further extend the applicability of the proposed variational family to the task of structured compression of convolutional neural network architecture. 
On MNIST dataset, we compress the architecture of LeNet5\footnote{\hyperlink{https://github.com/BVLC/caffe/tree/master/examples/mnist}{https://github.com/BVLC/caffe/tree/master/examples/mnist}}, which consists of 2 convolutional layers and 2 fully-connected layers.
We choose the model which has the best trainining cross-entropy during last 10 epochs.
After training, we plot a statistics (log of mode) of radius posteriors and prune it since it forms clearly separated clusters as shown in~\figref{lenet5_compression}.
In terms of various criteria for compressed architecture, compression using RDP shows well-balanced scores such as the number of parameters, FLOPs, and loss in accuracy.
We compute FLOPs for convolutional layers by $(K_w K_h C_{in} + 1)(I_h + P_h - K_h + 1)(I_w + P_w - K_w + 1) C_{out}$ where $I_h, I_w$ are input height and width, $K_h, K_w$ are kernel height and width, $P_w, P_h$ are padding height and width, and $C_{in}$, $C_{out}$ are the number of input and output channels. For fully-connected layers, we compute FLOPs by $(I_{in}+1)I_{out}$, where $I_{in}$ and $I_{out}$ are the number of input and output neurons, respectively.

The approach achieves competitive results with the state-of-the-art methods. 
As given in Table 2, the RDP architecture shows better compression for convolutional layers, which makes it score good at FLOPs.
The proposed pruning with a third of parameters of FLOPs as a Direct Optimization Objective(100K) (FDOO) is only slightly more computationally heavy. Similarly, RDP comes only second to BC-GHS in terms of parameter number but running with two-thirds of parameters of Bayesian Compression-Group Normal Jeffrey (BC-GNJ).  

\newcolumntype{P}[1]{>{\centering\arraybackslash}p{#1}}

\begin{table}[]
\vspace{-0.4em}
\label{tab:table_comp}
    \centering
    \begin{tabular}{|c|P{2cm}|c|c|c|}
        Method & \makecell{Pruned \\architecture} & FLOPs & Params & Error\\
  	\hline
  	
          \textbf{RDP} &\textbf{4-7-110-66}& \textbf{125K} & \textbf{20K} & \textbf{1.0\%} \\
        
          BC-GNJ & 8-13-88-13&  307K & 22K & 1.0\%\\
          BC-GHS & 5-10-76-16& 169K & \textbf{15K} & 1.0\%\\
          FDOO(100K) & 2-7-112-478& \textbf{119K} & 66K & 1.1\%\\
          FDOO(200K) & 3-8-128-499& 163K & 81K & 1.0\% \\
        
           GL & 3-12-192-500& 236K & 134K & 1.0\% \\
           GD & 7-13-208-16& 298K & 49K & 1.1\% \\
            SBP & 3-18-284-283& 295K & 164K & \textbf{0.9\%} \\

    \end{tabular}
    \vspace{-0.6em}
    \caption{The structured pruning of LeNet-5-Caffe with the original architecture 20-50-800-500. We benchmark our method against BC-GNJ, Bayesian Compression-Group HorseShoe(BC-GHS)~\cite{louizos2017bayesian}, FDOO~\cite{tang2018flops}, Generalized Dropout(GD)~\cite{srinivas2015data}, Group Lasso(GL)~\cite{wen2016learning}, 
    Structured Bayesian Pruning(SBP)~\cite{neklyudov2017structured}.
    }
    \vspace{-0.4em}
\end{table}

\section{Conclusion}\label{sec:conclusion}
We proposed a new variational family to capture dependencies between weight parameters in a structured way for Bayesian Neural Networks.
RDP's capability to capture the dependency between weights is empirically supported by its performance on regression tasks.
Also, RDP has a natural structure for compression and it scores better than other methods in multiple compression performance measures.

For practical implementations of variational inference with RDP, we proposed a simple approximation to the ratio of Bessel functions and a reparametrization trick for von-Mises Fisher(vMF) distribution. We expect our proposal to be impactful as there is a wide range of applications where variational inference with high dimensional vMF distributions could be useful. We currently use only the radius density for pruning. In future work, it would be intriguing to explore the directional component for further compression.



\newpage
\small
\bibliography{main}

\begin{thebibliography}{55}
\providecommand{\natexlab}[1]{#1}
\providecommand{\url}[1]{\texttt{#1}}
\expandafter\ifx\csname urlstyle\endcsname\relax
  \providecommand{\doi}[1]{doi: #1}\else
  \providecommand{\doi}{doi: \begingroup \urlstyle{rm}\Url}\fi

\bibitem[Abadi et~al.(2016)Abadi, Barham, Chen, Chen, Davis, Dean, Devin,
  Ghemawat, Irving, Isard, et~al.]{abadi2016tensorflow}
Abadi, M., Barham, P., Chen, J., Chen, Z., Davis, A., Dean, J., Devin, M.,
  Ghemawat, S., Irving, G., Isard, M., et~al.
\newblock Tensorflow: a system for large-scale machine learning.
\newblock In \emph{OSDI}, volume~16, pp.\  265--283, 2016.

\bibitem[Amos(1974)]{amos1974computation}
Amos, D.~E.
\newblock Computation of modified bessel functions and their ratios.
\newblock \emph{Mathematics of Computation}, 28\penalty0 (125):\penalty0
  239--251, 1974.

\bibitem[Banerjee et~al.(2005)Banerjee, Dhillon, Ghosh, and
  Sra]{banerjee2005clustering}
Banerjee, A., Dhillon, I.~S., Ghosh, J., and Sra, S.
\newblock Clustering on the unit hypersphere using von mises-fisher
  distributions.
\newblock \emph{Journal of Machine Learning Research}, 6\penalty0
  (Sep):\penalty0 1345--1382, 2005.

\bibitem[Batir(2008)]{batir2008inequalities}
Batir, N.
\newblock Inequalities for the gamma function.
\newblock \emph{Archiv der Mathematik}, 91\penalty0 (6):\penalty0 554--563,
  2008.

\bibitem[Bauckhage(2013)]{bauckhage2013computing}
Bauckhage, C.
\newblock Computing the kullback-leibler divergence between two weibull
  distributions.
\newblock \emph{arXiv preprint arXiv:1310.3713}, 2013.

\bibitem[Bauckhage(2014)]{bauckhage2014computing}
Bauckhage, C.
\newblock Computing the kullback-leibler divergence between two generalized
  gamma distributions.
\newblock \emph{arXiv preprint arXiv:1401.6853}, 2014.

\bibitem[Blundell et~al.(2015)Blundell, Cornebise, Kavukcuoglu, and
  Wierstra]{blundell2015weight}
Blundell, C., Cornebise, J., Kavukcuoglu, K., and Wierstra, D.
\newblock Weight uncertainty in neural networks.
\newblock \emph{arXiv preprint arXiv:1505.05424}, 2015.

\bibitem[Bui et~al.(2016)Bui, Hern{\'a}ndez-Lobato, Hernandez-Lobato, Li, and
  Turner]{bui2016deep}
Bui, T., Hern{\'a}ndez-Lobato, D., Hernandez-Lobato, J., Li, Y., and Turner, R.
\newblock Deep gaussian processes for regression using approximate expectation
  propagation.
\newblock In \emph{International Conference on Machine Learning}, pp.\
  1472--1481, 2016.

\bibitem[Damianou \& Lawrence(2013)Damianou and Lawrence]{damianou2013deep}
Damianou, A. and Lawrence, N.
\newblock Deep gaussian processes.
\newblock In \emph{Artificial Intelligence and Statistics}, pp.\  207--215,
  2013.

\bibitem[Davidson et~al.(2018)Davidson, Falorsi, De~Cao, Kipf, and
  Tomczak]{davidson2018hyperspherical}
Davidson, T.~R., Falorsi, L., De~Cao, N., Kipf, T., and Tomczak, J.~M.
\newblock Hyperspherical variational auto-encoders.
\newblock \emph{arXiv preprint arXiv:1804.00891}, 2018.

\bibitem[Dayan \& Hinton(1996)Dayan and Hinton]{dayan1996varieties}
Dayan, P. and Hinton, G.~E.
\newblock Varieties of helmholtz machine.
\newblock \emph{Neural Networks}, 9\penalty0 (8):\penalty0 1385--1403, 1996.

\bibitem[Figurnov et~al.(2018)Figurnov, Mohamed, and
  Mnih]{figurnov2018implicit}
Figurnov, M., Mohamed, S., and Mnih, A.
\newblock Implicit reparameterization gradients.
\newblock \emph{arXiv preprint arXiv:1805.08498}, 2018.

\bibitem[Gal \& Ghahramani(2016)Gal and Ghahramani]{gal2016dropout}
Gal, Y. and Ghahramani, Z.
\newblock Dropout as a bayesian approximation: Representing model uncertainty
  in deep learning.
\newblock In \emph{international conference on machine learning}, pp.\
  1050--1059, 2016.

\bibitem[Ghosh et~al.(2018)Ghosh, Yao, and Doshi-Velez]{pmlr-v80-ghosh18a}
Ghosh, S., Yao, J., and Doshi-Velez, F.
\newblock Structured variational learning of {B}ayesian neural networks with
  horseshoe priors.
\newblock In Dy, J. and Krause, A. (eds.), \emph{Proceedings of the 35th
  International Conference on Machine Learning}, volume~80 of \emph{Proceedings
  of Machine Learning Research}, pp.\  1744--1753, Stockholmsmässan, Stockholm
  Sweden, 10--15 Jul 2018. PMLR.
\newblock URL \url{http://proceedings.mlr.press/v80/ghosh18a.html}.

\bibitem[Graves(2011)]{graves2011practical}
Graves, A.
\newblock Practical variational inference for neural networks.
\newblock In \emph{Advances in neural information processing systems}, pp.\
  2348--2356, 2011.

\bibitem[Han et~al.(2015)Han, Mao, and Dally]{han2015deep}
Han, S., Mao, H., and Dally, W.~J.
\newblock Deep compression: Compressing deep neural networks with pruning,
  trained quantization and huffman coding.
\newblock \emph{arXiv preprint arXiv:1510.00149}, 2015.

\bibitem[Hassibi \& Stork(1993)Hassibi and Stork]{hassibi1993second}
Hassibi, B. and Stork, D.~G.
\newblock Second order derivatives for network pruning: Optimal brain surgeon.
\newblock In \emph{Advances in neural information processing systems}, pp.\
  164--171, 1993.

\bibitem[Hern{\'a}ndez-Lobato \& Adams(2015)Hern{\'a}ndez-Lobato and
  Adams]{hernandez2015probabilistic}
Hern{\'a}ndez-Lobato, J.~M. and Adams, R.
\newblock Probabilistic backpropagation for scalable learning of bayesian
  neural networks.
\newblock In \emph{International Conference on Machine Learning}, pp.\
  1861--1869, 2015.

\bibitem[Hoffman et~al.(2013)Hoffman, Blei, Wang, and
  Paisley]{hoffman2013stochastic}
Hoffman, M.~D., Blei, D.~M., Wang, C., and Paisley, J.
\newblock Stochastic variational inference.
\newblock \emph{The Journal of Machine Learning Research}, 14\penalty0
  (1):\penalty0 1303--1347, 2013.

\bibitem[Ifantis \& Siafarikas(1991)Ifantis and Siafarikas]{ifantis1991bounds}
Ifantis, E. and Siafarikas, P.
\newblock Bounds for modified bessel functions.
\newblock \emph{Rendiconti del Circolo Matematico di Palermo Series 2},
  40\penalty0 (3):\penalty0 347--356, 1991.

\bibitem[Jang et~al.(2016)Jang, Gu, and Poole]{jang2016categorical}
Jang, E., Gu, S., and Poole, B.
\newblock Categorical reparameterization with gumbel-softmax.
\newblock \emph{arXiv preprint arXiv:1611.01144}, 2016.

\bibitem[Kingma \& Ba(2014)Kingma and Ba]{kingma2014adam}
Kingma, D.~P. and Ba, J.
\newblock Adam: A method for stochastic optimization.
\newblock \emph{arXiv preprint arXiv:1412.6980}, 2014.

\bibitem[Kingma \& Welling(2013)Kingma and Welling]{kingma2013auto}
Kingma, D.~P. and Welling, M.
\newblock Auto-encoding variational bayes.
\newblock \emph{arXiv preprint arXiv:1312.6114}, 2013.

\bibitem[Kingma et~al.(2015)Kingma, Salimans, and
  Welling]{kingma2015variational}
Kingma, D.~P., Salimans, T., and Welling, M.
\newblock Variational dropout and the local reparameterization trick.
\newblock In \emph{Advances in Neural Information Processing Systems}, pp.\
  2575--2583, 2015.

\bibitem[Kreh(2012)]{kreh2012bessel}
Kreh, M.
\newblock Bessel functions.
\newblock \emph{Lecture Notes, Penn State-G{\"o}ttingen Summer School on Number
  Theory}, 82, 2012.

\bibitem[Lakshminarayanan et~al.(2017)Lakshminarayanan, Pritzel, and
  Blundell]{NIPS2017_7219}
Lakshminarayanan, B., Pritzel, A., and Blundell, C.
\newblock Simple and scalable predictive uncertainty estimation using deep
  ensembles.
\newblock In Guyon, I., Luxburg, U.~V., Bengio, S., Wallach, H., Fergus, R.,
  Vishwanathan, S., and Garnett, R. (eds.), \emph{Advances in Neural
  Information Processing Systems 30}, pp.\  6402--6413. Curran Associates,
  Inc., 2017.

\bibitem[Lebedev \& Lempitsky(2016)Lebedev and Lempitsky]{Lebedev_2016}
Lebedev, V. and Lempitsky, V.
\newblock Fast convnets using group-wise brain damage.
\newblock \emph{2016 IEEE Conference on Computer Vision and Pattern Recognition
  (CVPR)}, Jun 2016.
\newblock \doi{10.1109/cvpr.2016.280}.
\newblock URL \url{http://dx.doi.org/10.1109/CVPR.2016.280}.

\bibitem[Li et~al.(2016)Li, Kadav, Durdanovic, Samet, and Graf]{li2016pruning}
Li, H., Kadav, A., Durdanovic, I., Samet, H., and Graf, H.~P.
\newblock Pruning filters for efficient convnets.
\newblock \emph{arXiv preprint arXiv:1608.08710}, 2016.

\bibitem[Louizos \& Welling(2016)Louizos and Welling]{louizos2016structured}
Louizos, C. and Welling, M.
\newblock Structured and efficient variational deep learning with matrix
  gaussian posteriors.
\newblock In \emph{International Conference on Machine Learning}, pp.\
  1708--1716, 2016.

\bibitem[Louizos et~al.(2017)Louizos, Ullrich, and
  Welling]{louizos2017bayesian}
Louizos, C., Ullrich, K., and Welling, M.
\newblock Bayesian compression for deep learning.
\newblock In \emph{Advances in Neural Information Processing Systems}, pp.\
  3288--3298, 2017.

\bibitem[Luke(1972)]{luke1972inequalities}
Luke, Y.~L.
\newblock Inequalities for generalized hypergeometric functions.
\newblock \emph{Journal of Approximation Theory}, 5\penalty0 (1):\penalty0
  41--65, 1972.

\bibitem[MacKay(1995)]{mackay1995probable}
MacKay, D.~J.
\newblock Probable networks and plausible predictions—a review of practical
  bayesian methods for supervised neural networks.
\newblock \emph{Network: computation in neural systems}, 6\penalty0
  (3):\penalty0 469--505, 1995.

\bibitem[Maddison et~al.(2016)Maddison, Mnih, and Teh]{maddison2016concrete}
Maddison, C.~J., Mnih, A., and Teh, Y.~W.
\newblock The concrete distribution: A continuous relaxation of discrete random
  variables.
\newblock \emph{arXiv preprint arXiv:1611.00712}, 2016.

\bibitem[Mardia \& Jupp(2009)Mardia and Jupp]{mardia2009directional}
Mardia, K.~V. and Jupp, P.~E.
\newblock \emph{Directional statistics}, volume 494.
\newblock John Wiley \& Sons, 2009.

\bibitem[Molchanov et~al.(2017)Molchanov, Ashukha, and
  Vetrov]{molchanov2017variational}
Molchanov, D., Ashukha, A., and Vetrov, D.
\newblock Variational dropout sparsifies deep neural networks.
\newblock \emph{arXiv preprint arXiv:1701.05369}, 2017.

\bibitem[Naesseth et~al.(2016)Naesseth, Ruiz, Linderman, and
  Blei]{naesseth2016reparameterization}
Naesseth, C.~A., Ruiz, F.~J., Linderman, S.~W., and Blei, D.~M.
\newblock Reparameterization gradients through acceptance-rejection sampling
  algorithms.
\newblock \emph{arXiv preprint arXiv:1610.05683}, 2016.

\bibitem[Neal(2012)]{neal2012bayesian}
Neal, R.~M.
\newblock \emph{Bayesian learning for neural networks}, volume 118.
\newblock Springer Science \& Business Media, 2012.

\bibitem[Neklyudov et~al.(2017)Neklyudov, Molchanov, Ashukha, and
  Vetrov]{neklyudov2017structured}
Neklyudov, K., Molchanov, D., Ashukha, A., and Vetrov, D.~P.
\newblock Structured bayesian pruning via log-normal multiplicative noise.
\newblock In \emph{Advances in Neural Information Processing Systems}, pp.\
  6775--6784, 2017.

\bibitem[Neville et~al.(2014)Neville, Ormerod, Wand, et~al.]{neville2014mean}
Neville, S.~E., Ormerod, J.~T., Wand, M., et~al.
\newblock Mean field variational bayes for continuous sparse signal shrinkage:
  pitfalls and remedies.
\newblock \emph{Electronic Journal of Statistics}, 8\penalty0 (1):\penalty0
  1113--1151, 2014.

\bibitem[Paszke et~al.(2017)Paszke, Gross, Chintala, Chanan, Yang, DeVito, Lin,
  Desmaison, Antiga, and Lerer]{paszke2017automatic}
Paszke, A., Gross, S., Chintala, S., Chanan, G., Yang, E., DeVito, Z., Lin, Z.,
  Desmaison, A., Antiga, L., and Lerer, A.
\newblock Automatic differentiation in pytorch.
\newblock 2017.

\bibitem[Ritter et~al.(2018)Ritter, Botev, and Barber]{ritter2018a}
Ritter, H., Botev, A., and Barber, D.
\newblock A scalable laplace approximation for neural networks.
\newblock In \emph{International Conference on Learning Representations}, 2018.
\newblock URL \url{https://openreview.net/forum?id=Skdvd2xAZ}.

\bibitem[Ruiz-Antol{\'\i}n \& Segura(2016)Ruiz-Antol{\'\i}n and
  Segura]{ruiz2016new}
Ruiz-Antol{\'\i}n, D. and Segura, J.
\newblock A new type of sharp bounds for ratios of modified bessel functions.
\newblock \emph{Journal of Mathematical Analysis and Applications},
  443\penalty0 (2):\penalty0 1232--1246, 2016.

\bibitem[Slingo \& Palmer(2011)Slingo and Palmer]{slingo2011uncertainty}
Slingo, J. and Palmer, T.
\newblock Uncertainty in weather and climate prediction.
\newblock \emph{Phil. Trans. R. Soc. A}, 369\penalty0 (1956):\penalty0
  4751--4767, 2011.

\bibitem[Srinivas \& Babu(2015)Srinivas and Babu]{srinivas2015data}
Srinivas, S. and Babu, R.~V.
\newblock Data-free parameter pruning for deep neural networks.
\newblock \emph{arXiv preprint arXiv:1507.06149}, 2015.

\bibitem[Sun et~al.(2017{\natexlab{a}})Sun, Chen, and Carin]{pmlr-v54-sun17b}
Sun, S., Chen, C., and Carin, L.
\newblock {Learning Structured Weight Uncertainty in Bayesian Neural Networks}.
\newblock In Singh, A. and Zhu, J. (eds.), \emph{Proceedings of the 20th
  International Conference on Artificial Intelligence and Statistics},
  volume~54 of \emph{Proceedings of Machine Learning Research}, pp.\
  1283--1292, Fort Lauderdale, FL, USA, 20--22 Apr 2017{\natexlab{a}}. PMLR.
\newblock URL \url{http://proceedings.mlr.press/v54/sun17b.html}.

\bibitem[Sun et~al.(2017{\natexlab{b}})Sun, Chen, and Carin]{sun2017learning}
Sun, S., Chen, C., and Carin, L.
\newblock Learning structured weight uncertainty in bayesian neural networks.
\newblock In \emph{Artificial Intelligence and Statistics}, pp.\  1283--1292,
  2017{\natexlab{b}}.

\bibitem[Sun et~al.(2018)Sun, Zhang, Shi, and Grosse]{sun2018functional}
Sun, S., Zhang, G., Shi, J., and Grosse, R.
\newblock Functional variational bayesian neural networks.
\newblock 2018.

\bibitem[Tang et~al.(2018)Tang, Adhikari, and Lin]{tang2018flops}
Tang, R., Adhikari, A., and Lin, J.
\newblock Flops as a direct optimization objective for learning sparse neural
  networks.
\newblock \emph{arXiv preprint arXiv:1811.03060}, 2018.

\bibitem[Tucker et~al.(2017)Tucker, Mnih, Maddison, Lawson, and
  Sohl-Dickstein]{tucker2017rebar}
Tucker, G., Mnih, A., Maddison, C.~J., Lawson, J., and Sohl-Dickstein, J.
\newblock Rebar: Low-variance, unbiased gradient estimates for discrete latent
  variable models.
\newblock In \emph{Advances in Neural Information Processing Systems}, pp.\
  2627--2636, 2017.

\bibitem[Ullrich et~al.(2017)Ullrich, Meeds, and Welling]{ullrich2017soft}
Ullrich, K., Meeds, E., and Welling, M.
\newblock Soft weight-sharing for neural network compression.
\newblock \emph{arXiv preprint arXiv:1702.04008}, 2017.

\bibitem[Watson(1995)]{watson1995treatise}
Watson, G.~N.
\newblock \emph{A treatise on the theory of Bessel functions}.
\newblock Cambridge university press, 1995.

\bibitem[Wen et~al.(2016)Wen, Wu, Wang, Chen, and Li]{wen2016learning}
Wen, W., Wu, C., Wang, Y., Chen, Y., and Li, H.
\newblock Learning structured sparsity in deep neural networks.
\newblock In \emph{Advances in Neural Information Processing Systems}, pp.\
  2074--2082, 2016.

\bibitem[Yang \& Zheng(2017)Yang and Zheng]{yang2017sharp}
Yang, Z.-H. and Zheng, S.-Z.
\newblock Sharp bounds for the ratio of modified bessel functions.
\newblock \emph{Mediterranean Journal of Mathematics}, 14\penalty0
  (4):\penalty0 169, 2017.

\bibitem[Zhang et~al.(2017)Zhang, Sun, Duvenaud, and Grosse]{zhang2017noisy}
Zhang, G., Sun, S., Duvenaud, D., and Grosse, R.
\newblock Noisy natural gradient as variational inference.
\newblock \emph{arXiv preprint arXiv:1712.02390}, 2017.

\bibitem[Zhou et~al.(2016)Zhou, Alvarez, and Porikli]{zhou2016less}
Zhou, H., Alvarez, J.~M., and Porikli, F.
\newblock Less is more: Towards compact cnns.
\newblock In \emph{European Conference on Computer Vision}, pp.\  662--677.
  Springer, 2016.

\end{thebibliography}
\bibliographystyle{icml2019}
\onecolumn
\setcounter{section}{0}
\thispagestyle{empty}

\icmltitle{Supplementary Material\\Radial and Directional Posteriors for Bayesian Neural Networks}

\begin{icmlauthorlist}
\icmlauthor{Changyong Oh}{uva,intern}
\icmlauthor{Kamil Adamczewski}{mpi}
\icmlauthor{Mijung Park}{mpi}
\end{icmlauthorlist}

\printAffiliationsAndNotice{\icmlInternship}

\vspace{0.3in}
\section{Gradient of ELBO with von-Mises Fisher(vMF) distribution}\label{supp:repara_vMF}
\vspace{0.1in}
In variational inference, standard routine is to minimize negative evidence lower bound (ELBO)
\begin{equation}\label{eq:negative_elbo}
    \underbrace{-\Em_{q(\mW)}[p(\Dat \vert \mW)]}_{\text{Reconstruction error}} + \underbrace{D_{KL}(q(\mW) \Vert p(\mW))}_{\text{Regularizer}}
\end{equation}
where $\mW$ is a latent variable, $\Dat$ is data, $q(\mW)$ is a variational distribution on $\mW$, $p(\mW)$ is a prior on $\mW$ and $p(\Dat \vert \mW)$ is data likelihood given latent variable $\mW$.
\newline

Let's consider a von-Mises Fisher(vMF) distribution $q(\mW)=vMF(\mW \vert \mu, \kappa)$ on $\mathcal{S}^{m-1}$, which is reparametrizable \cite{davidson2018hyperspherical} through rejection sampling \cite{naesseth2016reparameterization}.

With reparametrization through rejection sampling, the gradient of reconstruction error in Eq.\eqref{negative_elbo} has the correction term \cite{naesseth2016reparameterization, davidson2018hyperspherical}
\begin{equation}\label{eq:vmf_gradient_correction}
    g_{cor} = -p(\Dat \vert \mW) \cdot \Big[-\frac{I_{m/2}(\kappa)}{I_{m/2-1}(\kappa)} + \nabla_{\kappa} \Big(\omega\kappa + \frac{1}{2}\log(1-\omega^2) + \log(\Big\lvert \frac{-2b}{(1-(1-\epsilon))^2} \Big\rvert ) \Big) \Big]
\end{equation}
where $I_{\nu}(\cdot)$ is modified Bessel function of the first kind with order $\nu$,
\begin{equation}
    b = \frac{-2\kappa + \sqrt{4\kappa^2 + (m-1)^2}}{m-1} \qquad \omega = h(\epsilon, \theta) = \frac{1-(1+b)\epsilon}{1-(1-b)\epsilon}
\end{equation}
$\mW$ is a vMF sample using reparametrization through rejection sampling, $\epsilon$ is a sample from the proposal distribution to reparametrize $\mW$ through rejection sampling. 
For more details, please refer to \cite{davidson2018hyperspherical}.


\vspace{0.3in}
\section{Instability of the calculation of the ratio of Bessel function $I_{\nu}(z)$}\label{supp:instable_bessel}
\vspace{0.1in}
Evaluating $I_{\nu}(z)$ is an expensive and unstable operation.
Moreover, evaluating the ratio $I_{\nu}(z)/I_{\nu-1}(z)$ is numerically more unstable for large $\nu$.

Let's first consider quantitatively how $I_{\nu}(z)$ behaves for large $\nu$. 
Thereby, we introduce necessary inequalities to bound $I_{\nu}(z)$ above.
\begin{prop}[\cite{luke1972inequalities}]\label{prop:bessel_bound}
    \begin{equation}
        1 < \Gamma(\nu+1)\Big(\frac{2}{z}\Big)^{\nu} I_{\nu}(z) < cosh(z) \quad \text{when} \quad z > 0, \quad \nu > 1/2
    \end{equation}
\end{prop}
\begin{prop}[\cite{batir2008inequalities}]\label{prop:gamma_bound}
    \begin{equation}
        \sqrt{2e} \Big( \frac{\nu+1/2}{e} \Big)^{\nu+1/2} < \Gamma(\nu+1) < \sqrt{2\pi} \Big( \frac{\nu+1/2}{e} \Big)^{\nu+1/2}
    \end{equation}
\end{prop}

From Proposition~\ref{prop:bessel_bound} and Proposition~\ref{prop:gamma_bound}, we have following upper bound of $I_{\nu}(z)$
\begin{equation}
    I_{\nu}(z) < \frac{1}{\Gamma(\nu+1)}\Big(\frac{z}{2}\Big)^{\nu} cosh(z) < \Big(\frac{z}{2}\Big)^{\nu} \frac{1}{\sqrt{2e}} \Big( \frac{e}{\nu+1/2} \Big)^{\nu + 1/2} \frac{e^z+e^{-z}}{2}
\end{equation}

For example, if
\begin{equation}
    z < \frac{\nu+1/2}{e}
\end{equation}
then
\begin{align}
    I_{\nu}(z) &<  \Big(\frac{z}{2}\Big)^{\nu} \frac{1}{\sqrt{2e}} \Big( \frac{e}{\nu+1/2} \Big)^{\nu + 1/2} \frac{e^z+e^{-z}}{2} \nonumber \\
    &< \Big( \frac{e}{\nu+1/2} \Big)^{1/2} \exp\Big(\frac{0.5}{e}\Big) \exp\Big(\nu(\frac{1}{e}-\log(2))\Big) \nonumber \\
    &= \Big( \frac{e}{\nu+1/2} \Big)^{1/2} \exp\Big(\frac{0.5}{e}\Big) \exp(-\varrho\nu)
\end{align}
where $\varrho=-(1/e-\log(2))=0.32526773938850295 > 0$.

Thus, for any finite $C > 0$ and any $\delta > 0$, we can find $\nu$ such that 
\begin{equation}
    I_{\nu}(z) < \delta \quad \forall z \in [0, C)
\end{equation}
Intuitively, a larger $\nu$ makes $I_{\nu}(z)$ arbitrarily small on longer intervals.

When it comes to the calculation of the ratio of Bessel functions with consecutive order, we often divide infinitesimally small number with infinitesimally small number where numerical instability is inevitable.

Even though exponentially scaled Bessel function $I_{\nu}(z) \times e^{-z}$ is recommneded when there is numerical issue with Bessel function $I_{\nu}(z)$, in the case of the ratio of Bessel functions, $I_{\nu}(z) \times e^{-z}$ only exacerbates this numerical issue (division between two infinitesimally small numbers) since $e^{-z} < 1$ for $z > 0$.

\vspace{0.3in}
\section{Bound on the ratio of modified Bessel function of the first kind $I_{\nu}(z)$}\label{supp:bessel_ratio_bound}
\vspace{0.1in}

Directly evaluating Bessel functions and calculating their ratio is expensive and unstable operation for large $\nu$.
However, their ratio between consecutive order enjoys very tight and algebraic bound
\begin{theorem}[Theorem 4 in \cite{ruiz2016new}]\label{thm:bessel_ratio_bound1}
    \begin{equation}
        \frac{z}{\nu-1/2 + \sqrt{(\nu + 1/2)^2 + z^2}} < \frac{I_{\nu}(z)}{I_{\nu-1}(z)} < \frac{z}{\nu-1 + \sqrt{(\nu + 1)^2 + z^2}} \quad \text{when} \quad \nu \ge 0
    \end{equation}
\end{theorem}
\begin{theorem}[Theorem 5 in \cite{ruiz2016new}]\label{thm:bessel_ratio_bound2}
    \begin{equation}
        B_2(\nu, z) < \frac{I_{\nu}(z)}{I_{\nu-1}(z)} < B_0(\nu, z) \quad \text{when} \quad \nu \ge 1/2
    \end{equation}
    where
    \begin{align}
        &B_{\alpha}(\nu, z) = \frac{z}{\delta_{\alpha}(\nu, z) + \sqrt{\delta_{\alpha}(\nu, z)^2 + z^2}} \nonumber \\
        &\delta_{\alpha}(\nu, z) = (\nu - 1/2) + \frac{\lambda}{2\sqrt{\lambda^2+z^2}} \qquad \lambda=\nu+(\alpha-1)/2 \nonumber
    \end{align}
\end{theorem}

\begin{figure}[t!]
\vspace{-0.1in}
\centerline{\includegraphics[width=1.0\textwidth]{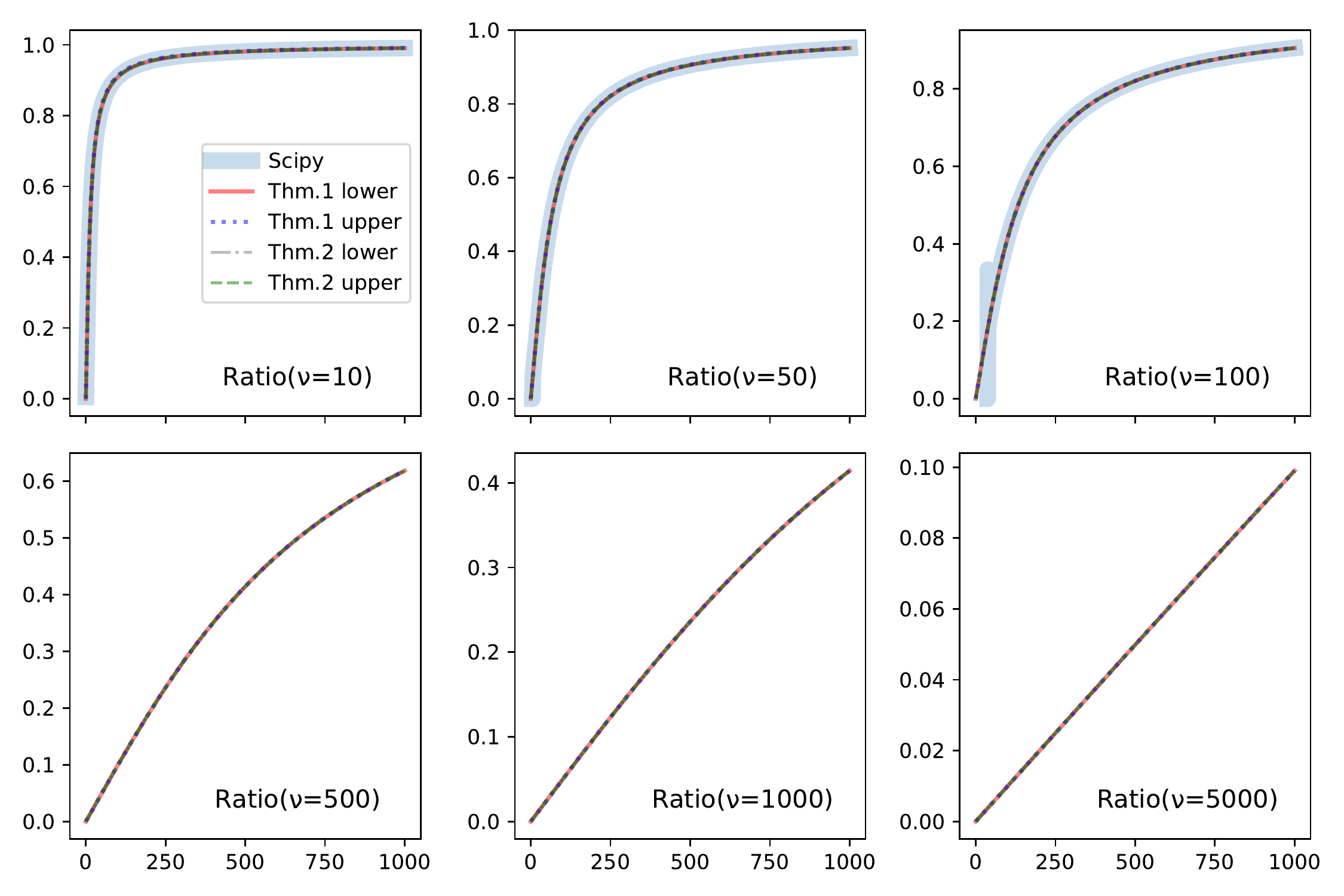}}
\caption{Calculation of $\frac{I_{\nu}(z)}{I_{\nu-1}(z)}$ with different methods (Thm.~\ref{thm:bessel_ratio_bound1}, Thm.~\ref{thm:bessel_ratio_bound2}, \textit{scipy.special.ive}). With Thm.~\ref{thm:bessel_ratio_bound1} and Thm.~\ref{thm:bessel_ratio_bound2}, the gap between lower bound and upper bound is negligible uniformly on $z$ and $\nu$. 
For $\nu \ge 500$, \textit{scipy.special.ive} cannot even generate a non-nan finite number. 
When $\nu = 100$, it exhibits numerical instability for $z < 100$.
}
\label{fig:bessel_ratio}
\end{figure}

In Fig~\ref{fig:bessel_ratio}, the quality of approximations using Thm.~\ref{thm:bessel_ratio_bound1} and Thm.~\ref{thm:bessel_ratio_bound2} is compared with the ratio using \textit{scipy.special.ive}.

There is also an algebraic bound with Amos-Type inequality for the ratio with smaller order in numerator as following form
\begin{equation}\label{eq:bessel_ratio_small_numer}
    z\frac{I_{\nu}(z)}{I_{\nu+1}(z)}
\end{equation}
in \cite{amos1974computation, yang2017sharp}.

However, we can use the relation
\begin{equation}\label{eq:bessel_ratio_relation}
    z\frac{I'_{\nu}(z)}{I_{\nu}(z)} = z\frac{I_{\nu-1}(z)}{I_{\nu}(z)} - \nu = z\frac{I_{\nu+1}(z)}{I_{\nu}(z)} + \nu
\end{equation}
derived from the recurrence relations \cite{watson1995treatise}
\begin{align}
    zI'_{\nu}(z) + \nu I_{\nu}(z) = z I_{\nu-1}(z) \nonumber \\
    zI'_{\nu}(z) - \nu I_{\nu}(z) = z I_{\nu+1}(z) \nonumber
\end{align}
and take advantage of tight bound of Thm.\ref{thm:bessel_ratio_bound1} or Thm.\ref{thm:bessel_ratio_bound2} to bound Eq.\ref{eq:bessel_ratio_small_numer}.

\vspace{0.3in}
\section{Estimation of gradient of ELBO with vMF using Bound on the ratio of Bessel function $I_{\nu}(z)$}\label{supp:gradient_approximation}
\vspace{0.1in}
There are several issues with calculating gradients of ELBO with vMF. 

As previously mentioned, the calculation of the ratio of modified Bessel function of the first kind $I_{\nu}(z)$ is extremely unstable for large $\nu$, which is suspected as one of reasons for degrading performance of Hyperspherical VAE~\cite{davidson2018hyperspherical} in high dimensions in addition to unintuitive behavior the area of hypersphere in high dimensions.

Also, in the most recent version of TensorFlow~\cite{abadi2016tensorflow} and PyTorch~\cite{paszke2017automatic} (TensorFlow r1.10 and PyTorch 0.4.1), modified Bessel function of the first kind $I_{\nu}(z)$ is not supported and CUDA implementation is not available yet.
Thus for efficient GPU computation, it may requires writing efficient CUDA code for modified Bessel function of the first kind.

Thanks to the tight and algebraic (rational function form) bound provided in Supp~\ref{supp:bessel_ratio_bound}, we can avoid calculation of $I_{\nu}(z)$ and have stable gradient estimation.

For the gradient correction for reconstruction error in Eq.\eqref{vmf_gradient_correction}, we approximate the ratio of Bessel functions as an average of lower bound and upper bound using Thm.\ref{thm:bessel_ratio_bound1} or Thm.\ref{thm:bessel_ratio_bound2}
\begin{align}
    \frac{I_{m/2}(\kappa)}{I_{m/2-1}(\kappa)} &\approx \frac{1}{2} \Big( \frac{z}{\nu-1/2 + \sqrt{(\nu + 1/2)^2 + z^2}} + \frac{z}{\nu-1 + \sqrt{(\nu + 1)^2 + z^2}} \Big) \\
    \frac{I_{m/2}(\kappa)}{I_{m/2-1}(\kappa)} &\approx \frac{B_2(\nu, z) + B_0(\nu, z)}{2}
\end{align}

For KL-divergence of location-agnostic prior for directional components, we use Thm.\ref{thm:bessel_ratio_bound1} or Thm.\ref{thm:bessel_ratio_bound2} 
\begin{align}
    \nabla_{\kappa_q}D_{KL}(vMF(\mu, \kappa_q) \Vert vMF(\mu, \kappa_p)) &= \frac{\partial}{\partial\kappa_q}(\kappa_q - \kappa_p) \frac{I_{m/2}(\kappa)}{I_{m/2-1}(\kappa)} + \frac{\partial}{\partial\kappa_q}\log(C_d(\kappa_q)) \nonumber \\
    =0.5(\kappa_q - \kappa_p) - &0.5(\kappa_q - \kappa_p) \frac{I_{m/2}(\kappa)}{I_{m/2-1}(\kappa)}\Big(\frac{I_{m/2+1}(\kappa)}{I_{m/2}(\kappa)} + \frac{I_{m/2-2}(\kappa)}{I_{m/2-1}(\kappa)} + \frac{I_{m/2}(\kappa)}{I_{m/2-1}(\kappa)} \Big)
\end{align}

Let $\rho_{\nu}^L$ and $\rho_{\nu}^U$ be lower bound and upper bound of the ratio $I_{\nu}(z)/I_{\nu-1}(z)$, respectively.
\begin{equation}
    \rho_{\nu}^L(z) < \frac{I_{\nu}(z)}{I_{\nu-1}(z)} < \rho_{\nu}^U(z)
\end{equation}

Then gradient KL-divergence is bounded as below
\begin{gather}
    \rho_{KL}^L < \nabla_{\kappa}D_{KL}(q(\mW) \Vert p(\mW)) < \rho_{KL}^U \nonumber \\
    \rho_{KL}^L = 0.5(\kappa_q - \kappa_p) - 0.5(\kappa_q - \kappa_p)\big[\rho_{m/2}^L (\rho_{m/2+1}^L + \frac{1}{\rho_{m/2-1}^U} + \rho_{m/2}^L)  \big]  \\
    \rho_{KL}^U = 0.5(\kappa_q - \kappa_p) - 0.5(\kappa_q - \kappa_p)\big[ \rho_{m/2}^U (\rho_{m/2+1}^U + \frac{1}{\rho_{m/2-1}^L} + \rho_{m/2}^U) \big]  \nonumber
\end{gather}



\vspace{0.3in}
\section{Approximate logarithm of modified Bessel function of the first kind $\log(I_{\nu}(z))$}\label{supp:log_bessel_approximation}
\vspace{0.1in}
We propose two approximations to the logarithm of modified Bessel function of the first kind $\log(I_{\nu}(z))$.

\subsection{Asymptotic bound}
We begin with tighter lower bound to $\log(I_{\nu}(z))$. 
\begin{prop}[Theorem 2.1 in \cite{ifantis1991bounds}]\label{prop:bessel_bound_better_lower}
    \begin{equation}
        cosh(z)^{\frac{1}{2(\nu+1)}} < \Gamma(\nu+1)\Big(\frac{2}{z}\Big)^{\nu} I_{\nu}(z) < cosh(z) \quad \text{when} \quad z > 0, \quad \nu > 1/2
    \end{equation}
\end{prop}
which provides tighter lower bound than Proposition~\ref{prop:bessel_bound}.

Simple algebra using Proposition~\ref{prop:bessel_bound_better_lower} gives us
\begin{equation}\label{eq:log_bessel_bound}
    \nu \log\Big(\frac{z}{2}\Big) - \log(\Gamma(\nu+1)) + \frac{1}{2(\nu+1)}\log(cosh(z)) < \log(I_{\nu}(z)) < \nu \log\Big(\frac{z}{2}\Big) - \log(\Gamma(\nu+1)) + cosh(z)
\end{equation}

By using
\begin{equation}
    z - \log(2) < \log(cosh(z)) < \log(e^z+1) - \log(2) \quad \text{when} \quad z > 0
\end{equation}
and  
\begin{equation}
    0 < \log(e^z + 1) - z < \log(2) \quad \text{when} \quad z > 0
\end{equation}
we can simplify bound in eq.~\eqref{log_bessel_bound} in more by approximating $\log(Cosh(z))$ with $z-\log(2)$.

We approximate logarithm of Bessel function with the average of the bounds
\begin{equation}
    \log(I_{\nu}(z)) \approx  \nu\log(z) + \eta z - (\eta + \nu)\log(2) - \log(\Gamma(\nu+1))
\end{equation}
where $\eta = \frac{\nu + 1/2}{2(\nu + 1)}$.
This approximation works well for small $z$, especially $z < \nu$.

For large $z \ge \nu$, we use approximate $\log(I_{\nu}(z))$ with its asymptotic behavior.

\begin{theorem}[Theorem 2.13 in \cite{kreh2012bessel}]
    \begin{equation}
        I_{\nu}(z) \sim \sqrt{\frac{1}{2\pi z}}e^z
    \end{equation}
\end{theorem}
We have an approximate for large $z \ge \nu$
\begin{equation}
    \log(I_{\nu}(z)) \approx z - 0.5\log(z) - 0.5\log(2\pi)
\end{equation}

Logarithm of modified Bessel function of the first kind $\log(I_{\nu}(z))$ is approximated by
\begin{equation}
    \log(I_{\nu}(z)) \approx
    \begin{cases}
        \nu\log(z) + \eta z - (\eta + \nu)\log(2) - \log(\Gamma(\nu+1)) & z < \nu \\
        z - 0.5\log(z) - 0.5\log(2\pi) & z \ge \nu
    \end{cases}
\end{equation}
where $\eta = \frac{\nu + 1/2}{2(\nu + 1)}$.

\subsection{Log Ratio Series}

\begin{figure}[t!]
\vspace{-0.1in}
\centerline{\includegraphics[width=1.0\textwidth]{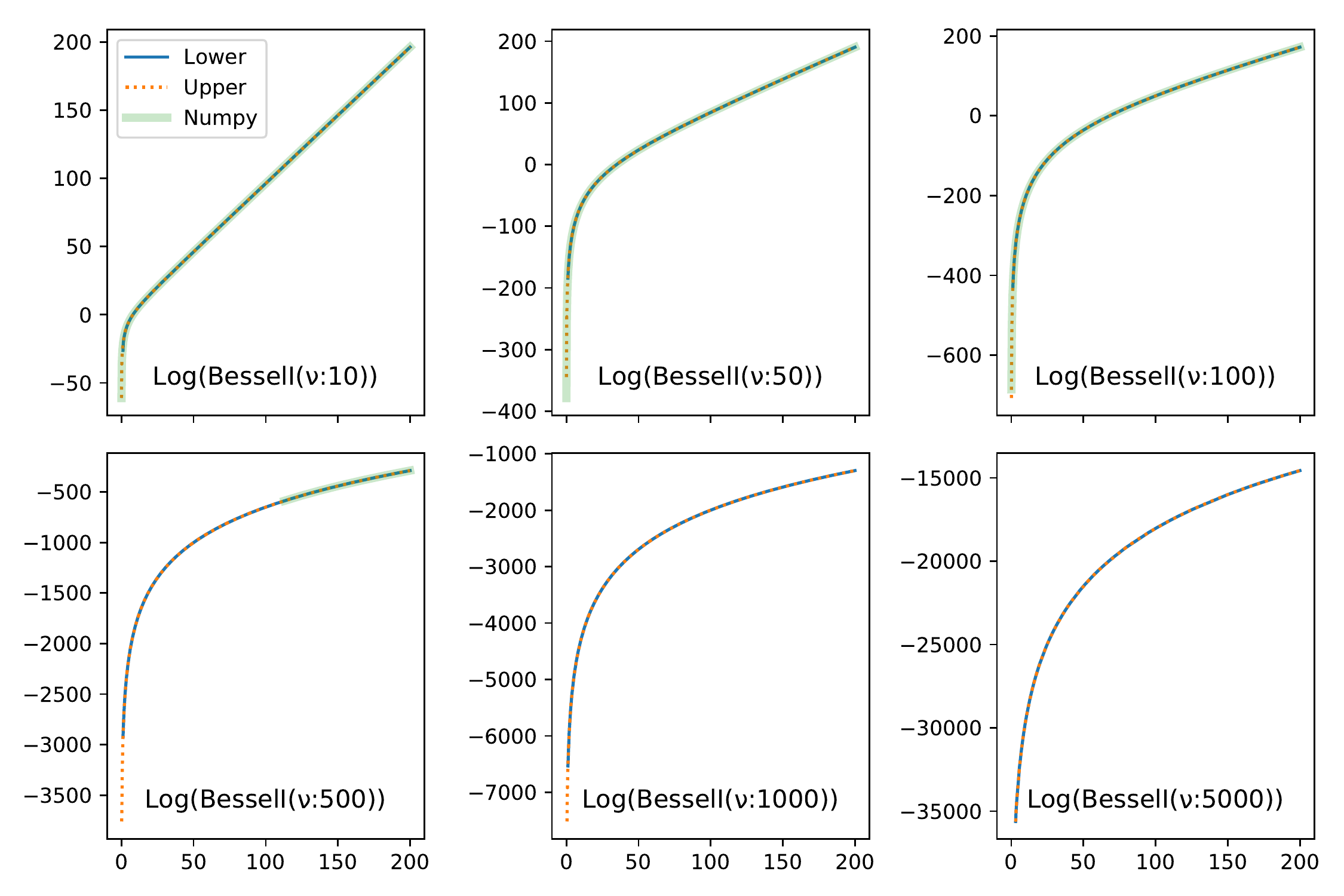}}
\caption{Calculation of $\log(I_{\nu}(z))$ with different methods (Thm.~\ref{thm:bessel_ratio_bound2}, \textit{scipy.special.ive}). 
Approximation using Thm.~\ref{thm:bessel_ratio_bound2} gives stable result uniformly on $z$ and $\nu$. 
For $\nu \ge 500$, \textit{scipy.special.ive} cannot even generate a non-nan finite number when $z < 100$.
}
\label{fig:log_bessel}
\end{figure}

We can make full use of tight bound on the ratio in Supp.~\ref{supp:bessel_ratio_bound} by using below relation
\begin{equation}
    \log(I_{\nu}(z)) = \log(I_{\nu-\lfloor \nu \rfloor}(z)) + \sum_{n=0}^{\lfloor \nu-1 \rfloor} \log\Big(\frac{I_{\nu-n}(z)}{I_{\nu-1-n}(z)}\Big)
\end{equation}
Since the bounds on the ratio is tight and stable and for small $\nu$, $I_{\nu}(z)$ can be calculated stably.
For large $\nu$, this approximation is still stable but slow.

The quality of this approximation using Thm.~\ref{thm:bessel_ratio_bound2} is given in Fig~\ref{fig:log_bessel}. As observed in Fig.~\ref{fig:log_bessel}, the method approximates tightly but at certain computational cost (taking twice than the method using \textit{scipy.special.ive}).

This approximation can be used in forward pass of KL-divergence with vMF in \eqref{negative_elbo}. 
Since, in our implementation of vMF KL-divergence, backward pass is calculated without using outputs of forward pass, even using arbitrary value for forward pass of vMF KL-divergence does not influence learning.
However, this approximation which is much cheaper than Bessel function can be used as an estimate of ELBO to monitor training progress.

\vspace{0.3in}
\section{Kullback-Leibler Divergence}\label{supp:kl_divergence}
\vspace{0.1in}

\subsection{von-Mises Fisher $q \sim vMF(\mu_q, \kappa_q)$ and $p \sim vMF(\mu_p, \kappa_p)$}
\begin{equation}
    D_{KL}(vMF(\mu_q, \kappa_q) \Vert vMF(\mu_p, \kappa_p)) = (\kappa_q - \kappa_p\mu_p^T\mu_q) \frac{I_{m/2}(\kappa_q)}{I_{m/2-1}(\kappa_q)}  + \log(C_m(\kappa_q)) - \log(C_m(\kappa_p))
\end{equation}

The gradient with respect to $\mu_q$ and $\kappa_q$ is given as follows

\begin{align}
    &\nabla_{\kappa_q}D_{KL}(vMF(\mu_q, \kappa_q) \Vert vMF(\mu_p, \kappa_p)) \nonumber \\
    &= \frac{1}{2}(\kappa_q - \kappa_p\mu_p^T\mu_q)\Big[ \frac{I_{m/2+1}(\kappa_q)}{I_{m/2-1}(\kappa_q)} - \frac{I_{m/2}(\kappa_q)}{I_{m/2-1}(\kappa_q)} \Big(\frac{I_{m/2-2}(\kappa_q)}{I_{m/2-1}(\kappa_q)} + \frac{I_{m/2}(\kappa_q)}{I_{m/2-1}(\kappa_q)} \Big) + 1 \Big] \nonumber \\
    &= \frac{1}{2}(\kappa_q - \kappa_p\mu_p^T\mu_q)  \Big[ \frac{I_{m/2}(\kappa_q)}{I_{m/2-1}(\kappa_q)} \Big(\frac{I_{m/2+1}(\kappa_q)}{I_{m/2}(\kappa_q)} - 2 \frac{I_{m/2}(\kappa_q)}{I_{m/2-1}(\kappa_q)} - \frac{m - 2}{\kappa_q} \Big) + 1 \Big]
\end{align}
using $\nabla_{\kappa_q}\log(C_m(\kappa_q)) = -\frac{I_{m/2}(\kappa_q)}{I_{m/2-1}(\kappa_q)}$
\begin{equation}
    \nabla_{\mu_q}D_{KL}(vMF(\mu_q, \kappa_q) \Vert vMF(\mu_p, \kappa_p))
    = - \kappa_p\mu_p \frac{I_{m/2}(\kappa_q)}{I_{m/2-1}(\kappa_q)}
\end{equation}

\subsection{$q \sim \Gamma(\alpha_q, \beta_q)$ and $p \sim \Gamma(\alpha_p, \beta_p)$ \cite{bauckhage2014computing}}
\begin{equation}
    (\alpha_q - \alpha_p)\Psi(\alpha_q) - \log(\mathit{\Gamma}(\alpha_q)) + \log(\mathit{\Gamma}(\alpha_p)) + \alpha_p (\log(\beta_q) - \log(\beta_p)) + \alpha_q\frac{\beta_p - \beta_q}{\beta_q}
\end{equation}
where $\mathit{\Gamma}$ is gamma function and $\Psi$ is digamma function.

\subsection{$q \sim \mathcal{W}(\lambda_q, k_q)$ and $p \sim \mathcal{W}(\lambda_p, k_p)$ \cite{bauckhage2013computing}}
\begin{equation}
    (\log(k_q) - k_q\log(\lambda_q)) - (\log(k_p) - k_p\log(\lambda_p)) + (k_q - k_p)\Big(\log(\lambda_q - \frac{\gamma}{k_q})\Big) + \Big(\frac{\lambda_q}{\lambda_p}\Big)^{k_p} \mathit{\Gamma}\Big(\frac{k_q}{k_p}+1\Big)-1
\end{equation}
where $\gamma = 0.5772$ is  Euler-Mascheroni constant and $\mathit{\Gamma}$ is gamma function.

\subsection{$q \sim \mathcal{LN}(\mu, \sigma^2)$ and $p \sim \Gamma(\alpha, \beta)$ \cite{louizos2017bayesian}}
\begin{equation}
    \alpha\log(\beta) + \log(\mathit{\Gamma}(\alpha)) - \alpha \mu + \frac{1}{\beta}\exp(-\mu + \sigma^2/2) - (\log(2\sigma^2) + 1)/2
\end{equation}
where $\mathit{\Gamma}$ is gamma function.

\subsection{$q \sim \mathcal{LN}(\mu, \sigma^2)$ and $p \sim \iota\Gamma(\alpha, \beta)$ \cite{louizos2017bayesian}}
\begin{equation}
    -\alpha\log(\beta) + \log(\mathit{\Gamma}(\alpha)) + \alpha \mu + \frac{1}{\beta}\exp(\mu + \sigma^2/2) - (\log(2\sigma^2) + 1)/2
\end{equation}
where $\mathit{\Gamma}$ is gamma function.

\end{document}